%% file: main.tex
\documentclass[letterpaper, 10 pt, conference]{ieeeconf}  
\IEEEoverridecommandlockouts

\usepackage{amsmath}
\usepackage{algorithm}
\usepackage{algorithmic}
\usepackage{wrapfig}
\usepackage{todonotes}
\usepackage{soul}

\newcommand{\comment}[1]{}

\begin{document}





\title{\LARGE \bf 
Self-Supervised Learning of State Estimation for Manipulating Deformable Linear Objects}

\author{Mengyuan Yan$^{1}$, Yilin Zhu$^{1}$, Ning Jin$^{2}$, Jeannette Bohg$^{1}$
\thanks{$^{1}$Mengyuan Yan, Yilin Zhu, and Jeannette Bohg are with School of Engineering, Stanford University.
        {\tt\small \{mengyuan, ylzhu, bohg\}@stanford.edu}}%
\thanks{$^{2}$Ning Jin is with Calico Labs. This research is done during Ning's PhD at Stanford University. 
        {\tt\small jennyjin@calicolabs.com}}%
\thanks{The Okawa Foundation and Toyota Research Institute ("TRI")  provided funds to assist the authors with their research but this article solely reflects the opinions and conclusions of its authors and not of the Okawa Foundation, TRI or any other Toyota entity.}
}

\maketitle
\thispagestyle{empty}
\pagestyle{empty}

\begin{abstract}
We demonstrate model-based, visual robot manipulation of linear deformable objects. Our approach is based on a state-space representation of the physical system that the robot aims to control. This choice has multiple advantages, including the ease of incorporating physics priors in the dynamics model and perception model, and the ease of planning manipulation actions. 
In addition, physical states can naturally represent object instances of different appearances. Therefore, dynamics in the state space can be learned in one setting and directly used in other visually different settings. This is in contrast to dynamics learned in pixel space or latent space, where generalization to visual differences are not guaranteed. Challenges in taking the state-space approach are the estimation of the high-dimensional state of a deformable object from raw images, where annotations are very expensive on real data, and finding a dynamics model that is both accurate, generalizable, and efficient to compute. We are the first to demonstrate self-supervised training of rope state estimation on real images, without requiring expensive annotations. This is achieved by our novel self-supervising learning objective, which is generalizable across a wide range of visual appearances. With estimated rope states, we train a fast and differentiable neural network dynamics model that encodes the physics of mass-spring systems. Our method has a higher accuracy in predicting future states compared to models that do not involve explicit state estimation and do not use any physics prior, while only using 3\% of training data. We also show that our approach achieves more efficient manipulation, both in simulation and on a real robot, when used within a model predictive controller.
\end{abstract}


\section{Introduction}
\input{introduction.tex}

\section{Related Work}

\input{related.tex}

\section{Method}
\input{method.tex}

\section{Experiments and Results}
\input{result.tex}

\section{Conclusion}
\input{conclusion.tex}

\vspace{-5pt}
\section*{Acknowledgments}
We thank Professor Ronald Fedkiw for useful discussions on deformable object simulation.

\vspace{-5pt}
\bibliographystyle{IEEEtran}
\bibliography{IEEEabrv,ref}
\end{document}


\maketitle

\section{Perception Model}
\begin{figure}[h]
    \centering
    \includegraphics[width=\linewidth,trim={0.7in 0in 2.3in 0in},clip]{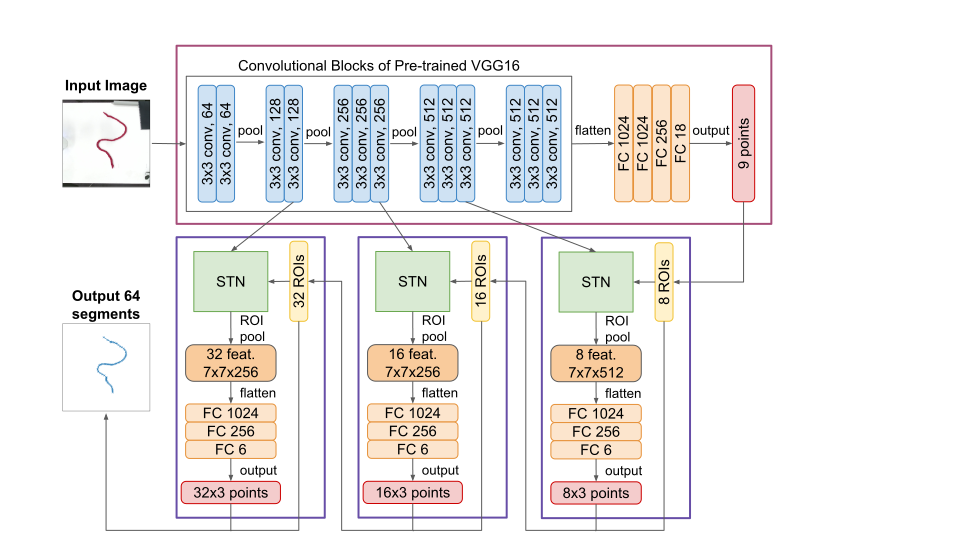}
    \caption{Detailed architecture of our perception network. 
    Top: Initial coarse prediction of 8 segments on the rope. We first run input images through the convolutional blocks of a pretrained VGG16 network to extract feature maps at different resolutions, and the feature map from the last convolutional layer then goes through 4 fully connected layers to output initial estimation for the 9 points that define the 8 ROIs. 
    Bottom: Hierarchical refinement of the estimation via spatial transformer networks. Starting from the initial coarse prediction from above, we iteratively refine the estimation, where each time we double the number of ROIs. We run a total of $3$ steps, and end up with $64$ segments as our final output estimation.
    }
    \label{fig:detailed_perception_arch}
\end{figure}

In this section, we describe the details of our perception model, which can be broken down into two components: 1) an initial coarse estimation component using a pretrained VGG16~\cite{vgg} network and a few fully connected layers (Fig.~\ref{fig:detailed_perception_arch} top) and 2) a hierarchical refinement component using spatial transformer networks on feature maps from the VGG network (Fig.~\ref{fig:detailed_perception_arch} bottom). 

As shown in Fig.~\ref{fig:detailed_perception_arch} top, the initial coarse estimation takes in $224\times224$ images and outputs $9$ points (18 numbers in $2D$) that define 8 segments to approximate the shape of the rope. 
We first run the image through the convolutional blocks of a pretrained VGG network to extract feature maps at different resolutions.
Then the feature map from the last convolutional layer goes through 4 fully connected layers of size $1024, 1024, 256$, and $18$ to output prediction for the 9 points. 
During training, the VGG network is frozen and only the fully connected layers are updated. 

The initial estimation is then refined hierarchically as shown in Fig.~\ref{fig:detailed_perception_arch} bottom. 
Starting from the initial $8$ segments, we repeatedly refine the estimation by doubling the number of ROIs each time, i.e., $8\rightarrow16\rightarrow32\rightarrow64$. At step $i$ ($i=1,2,\dots$), we start with $8\times 2^{i-1}$ ROIs, and for each ROI, we crop out feature maps from the last convolutional layer in the $(5-i)$-th convolutional block of the pretrained VGG network, and transform the cropped feature maps into $7\times7$ spatial resolution using the STN. 
The features are then flattened and go through 3 more fully connected layers of size $1024, 256$, and $6$ to predict $(8\times 2^{i-1})\times3$ points, where each triplet of points is predicted independently from the corresponding feature, and refines the corresponding segment by estimating its start-point, mid-point, and end-point. These estimations are then combined, where end-points of previous ROI and start-points of the next ROIs are averaged, and mid-points are kept as estimated. After step $i$, we get $8\times 2^i$ ROIs for the next step. 
We run a total of $3$ steps, and end up with $64$ segments as our final output for the rope state estimation.

\comment{
\section{Image loss comparison}
In this section, we provide more mathematical insights into why our proposed image loss is more generalizable than L2 loss, where the rendered grayscale image is colored by the mean of each of the two clusters. We show that optimizing with the L2 loss is closely related to K-means clustering in RGB space, and many assumptions of K-means clustering is violated on real images.

Using the same notation as in the paper, where $x_i$ are the RGB values for each pixel, $\mu_1$ and $\mu_2$ are the mean for the rope RGB cluster and the background RGB cluster respectively, $w_{ik}$ is the membership weight from pixel $i$ to cluster $k$. the L2 loss can be written as
\begin{equation}
    \sum_i (x_i - w_{i1}\mu_1 - w_{i2}\mu_2)^2. 
\end{equation}
For the case of binary membership weights, i.e., $w_{ik} \in {0,1}$, this loss is equivalent to K-means with 2 clusters
\begin{equation}
    \sum_i w_{i1}(x_i - \mu_1)^2 + w_{i2}(x_i - \mu_2)^2. 
\end{equation}
Therefore, intuitively, when optimizing the rope state using the L2 loss, we are using K-means to cluster the RGB space distribution into 2 clusters, and updating the rope state to be consistent with the resulting color-based segmentation. K-means clustering has the following assumptions: (1) The variance of each cluster is roughly equal. (2) The number of data points in each cluster is roughly equal. (3) The distribution of each cluster is roughly isotropic (spherical). All of these assumptions are broken in our case for real images. Assumption (1) is broken when the rope is textured with two colors. Assumption (2) is broken, since the background cluster have significantly more pixels than the rope cluster. Assumption (3) is also broken, because for real images, often the variance of brightness of pixels is much larger than the variance of hue, due to lighting and shadows. 

On the other hand, our proposed loss is closely related to using Gaussian mixture model (GMM) to cluster the RGB space. Using the Gaussian mixture model does not have any of the above assumptions, thus is more generalizable on real images.
}

\begin{algorithm}[!t]
\caption{Algorithm for finetuning perception network.}
\label{alg:finetune}
\small
\begin{algorithmic}

\STATE Dataset consists of pairs of neighboring RGB images.
\STATE hyper-parameters: $loss\_thresh$, batch size=48
\FOR {$epoch = 1,\dots,N$}
    \STATE Shuffle dataset
    \IF{$num\_trained \text{ did not increase}$}
        \STATE $ loss\_thresh \gets 0.98~loss\_thresh $
    \ENDIF
    \STATE $num\_trained \gets 0$
    \FOR {each batch}
        \STATE {$I_i, i=0,\dots, 47$ are RGB images.}
        \STATE {$I_{2j}$ and $I_{2j+1}$ are neighboring images.}
        \STATE Forward pass $S_i = \text{Net}(I_i)$
        \STATE Evaluate loss $L_i = \text{ImageLoss}(S_i, I_i)$
        \STATE Total training loss $L \gets 0$

        \FOR {$i=0,\dots, 47$}
            \IF {$L_i < loss\_thresh$}
            \STATE $L = L + L_i$
            \STATE $num\_train = num\_train + 1$
            \ENDIF
        \ENDFOR
        \FOR{$j=0,\dots,23$}
            \IF{$L_{2j} < loss\_thresh$ and $L_{2j+1} > loss\_thresh$}
                \STATE $L = L + l2(\text{StopGradient}(S_{2j})-S_{2j+1})$
            \ENDIF
            \IF{$L_{2j} > loss\_thresh$ and $L_{2j+1} < loss\_thresh$}
                \STATE $L = L + l2(\text{StopGradient}(S_{2j+1})-S_{2j})$
            \ENDIF
        \ENDFOR
    \STATE{Update network with gradients of $L$.}
    \ENDFOR
\ENDFOR

\end{algorithmic}
\end{algorithm}

\section{Self-supervised Network Finetuning}
The details of the finetuning algorithm is summarized in Algorithm \ref{alg:finetune}. We optimize the network weights with Adam optimizer to minimize the self-supervising learning objective (image loss). To incorporate automatic curriculum learning, we set a loss threshold, which is always a negative number. All training samples that have a smaller loss than the threshold are taken into account for the gradient updates, and other samples are omitted. Experimentally, we set the loss threshold to include about top $10\%$ training samples in the first epoch. If the number of training samples below the threshold stops growing for 2 epochs, we increase the threshold by multiplying $0.98$.

To incorporate temporal consistency, we restructure the dataset so that it consists of pairs of neighboring RGB images. During training, for each pair of images, if one of them has image loss below the threshold and the other has image loss above the threshold, the worse one is trained to minimize the L2 loss with the prediction of the better one as its "ground truth label", and the better one is trained to minimize the image loss. If both images have image loss below the threshold, then both images are trained with image loss. If both images have image loss above the threshold, they are omitted for this epoch.

\section{Neural Network Dynamics Model}
In this section we give the equations that describe our bi-directional LSTM model. For each of the 65 nodes on the rope, its input contains its position $p_i\in\mathbbm{R}^2$, action $a_i\in\mathbbm{R}^2$ (which is 0 unless the action applies on this node), and an indicator $f_i=\mathbbm{1}(|a_i|>0)$. The inputs are concatenated into $x_i=(p_i,a_i,f_i)\in\mathbbm{R}^5$ for $i=1,\dots,65$. $p_i$ are retrieved from simulation during training and estimated from images during evaluation.

The bi-directional LSTM is constructed as
\begin{align*}
    h^L_1 &= 0, \\
    z^L_i, h^L_{i+1} &= LSTM(x_i, h^L_i), i=1,\dots,65, \\
    h^R_{65} &= 0, \\
    z^R_i, h^R_{i-1} &= LSTM(x_i, h^R_i), i=65,\dots,1, \\
    y_i &= w_L z^L_i + w_R z^R_i + w_I x_i, 
\end{align*}
where the superscript $L$ denotes the LSTM propagating from node $1$ to node $65$, and the superscript $R$ denotes the LSTM in the reverse direction. $h^L_i$ and $h^R_i$ are the memory units, $z^L_i$ and $z^R_i$ are the output units. The LSTM cell has one layer with $256$ units and ReLu6 activation. LSTM outputs $z^R_i$ and $z^L_i$ are concatenated together with the input $x_i$, and fed into one more linear layer to predict the position $p_i^{\text{out}}$ of the $i$th node after the action is performed. The LSTM can be applied repeatedly on a sequence of actions for long-horizon prediction.

\section{Heuristics Used For Action Planning}
In this section we describe the heuristics used to generate candidate action sequences for MPPI in our manipulation experiments. These heuristics are made possible because we explicitly estimate rope states for the current observation and the goal image.

At each time step $t$, with current rope state $s_t = \{p_{i,t}, i=1,\dots,65$\}, and goal state $s_{\goal} = \{ p_{i,\goal}, i=1,\dots,65\}$, we densely sample every other node, i.e. node $1,3,5,\dots,65$ as candidate grasping points. Then, for each candidate grasping point, we generate $30$ sequences each containing $10$ displacement vectors. To generate each sequence starting at point $i$, we first calculate the unit vector $e_1$ in the direction of $p_{i,\goal}-p_{i,t}$, and $e_2$ such that $e_2 \bot e_1$. all actions are initialized as $0.8 a_{\max} e_1$, where $a_{\max}$ is the maximum magnitude of displacement vectors. In case $d_{i,t}=|p_{i,\goal}-p_{i,t}| < 8 a_{\max}$, i.e., the goal position can be reached before the $10$ action finishes, the final actions are clipped in magnitude so that the sequence should bring the grasping point to $p_{i,\goal}$. Then we add exploration noises to the sequence. Exploration noise is specified by three random variables $\delta_{x}$, $\delta_{y}$ and $\delta_{c}$. Intuitively, $\delta_{x}$ and $\delta_{y}$ moves the endpoint of the $10$-action trajectory in the 2D plane, while $\delta_{c}$ modifies the trajectory from a straight line to a curve, without changing the endpoint. Mathematically, the $10$ displacement vectors $a_{i,1}, a_{i,2}, \dots, a_{i,10}$ are modified as
\begin{equation}
    a_{i,t} = a_{i,t}^{init} + (\delta_x, \delta_y) + \delta_c cos(t\pi/10) e_2.
\end{equation}

\comment{
\section{Rope tracking during real robot manipulation}
As discussed in the main paper, due to the limited size of the real-world dataset and the very large state space of ropes, the network does not generalize well to rope states outside of the training distribution. Experimentally, we observe that most of the overfitting is from the coarsest prediction layer, and the refinement layers generalize well if given a reasonable coarse prediction. This motivates us to further exploit temporal information during manipulation. 

\begin{figure}
    \centering
    \includegraphics[width=0.45\textwidth]{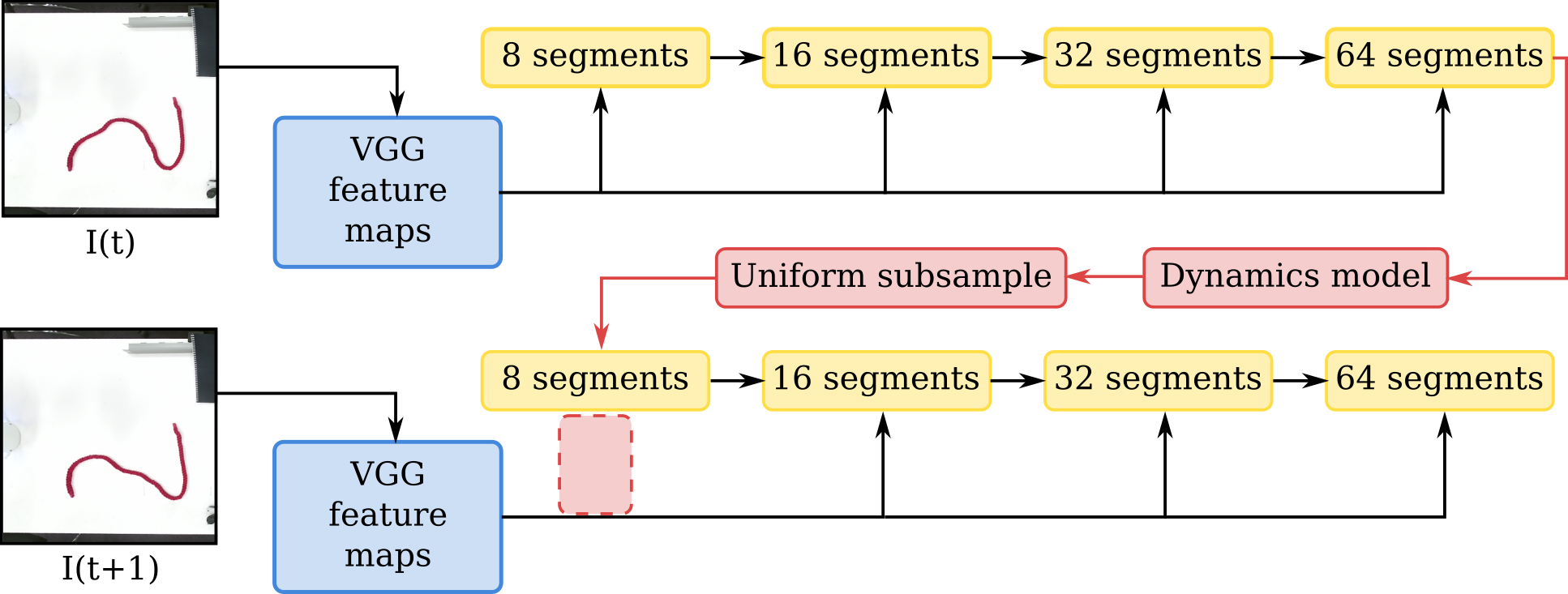}
    \caption{Illustration of rope tracking by recurrently using dynamics model prediction to replace the coarsest level rope state prediction. Note the added path in red that passes information from previous frame to the next frame, replacing the dashed red rectangle which would independently estimate the rope state for each frame.}
    \label{fig:tracking}
\end{figure}

As illustrated in Fig. ~\ref{fig:tracking}, the perception model estimates the latest rope state $s_t$ from image $I_t$. After MPC plans the optimal action $a_t$, we use the learned dynamics model to predict the next state $\hat{s}_{t+1}$. $\hat{s}_{t+1}$ is coarsened by reducing the $64$ segments to $8$ segments through merging of each group of $8$ consecutive segments into $1$. The resulting $8$ coarse segments are fed into the CNN's first STN, together with the next image $I_{t+1}$. Thus instead of estimating the $8$ segments at the coarsest level from image $I_{t+1}$, they are assumed to be equal the predicted segments from the previous frame. The CNN only refines the coarsened $\hat{s}_{t+1}$ into the next estimated state $s_{t+1}$.

\jean{Up till here, I totally think you should keep everything. But I'm not sure if you want to keep the EKF part because it is a little unclear right now. If you keep it, don't say that this is an approximation of an EKF. Make it weaker and say, that there are parallels between the approaches or that there are similarities.}
This recurrent process is an approximation to an \emph{Extended Kalman Filter} (EKF). In an EKF, the process model is $s_{t+1} = f(s_t, a_t) + w_t$, and the observation model is $x_t = h(s_t) + v_t$. Assuming that the current state estimate of the system is normally distributed with mean $s_t$ and covariance $\Sigma_t$, the EKF update rules are:

\begin{equation*}
    \begin{array}{rcl}
        \hat{s}_{t+1} &=& f(s_t, a_t)  \\
        \hat{\Sigma}_{t+1} &=& F_t \Sigma_t F_t^T + Q_t \\
        \Delta &=& x_{t+1}-h(\hat{s}_{t+1}) \\
        S_{t+1} &=& H_{t+1} \hat{\Sigma}_{t+1} H_{t+1}^T + R_{t+1} \\
        s_{t+1} &=& \hat{s}_{t+1} + \hat{\Sigma}_{t+1} H_{t+1}^T S_{t+1}^{-1} \Delta \\
        \Sigma_{t+1} &=& (I - \hat{\Sigma}_{t+1} H_{t+1}^T S_{t+1}^{-1} H_{t+1}) \hat{\Sigma}_{t+1}
    \end{array}
\end{equation*}
where $Q_t$ and $R_t$ are the covariance matrices for the process noise $w_t$ and observation noise $v_t$, $F_t$ and $H_t$ are the Jacobians of the functions $f$ and $h$ with respect to the current state $s_t$ and action $a_t$. 
\jean{Actually, shouldn't there be a second Jacobian that linearizes $f$ around the current action and appear in the second line as an additional term on the right side?}
\jean{Since you think of $s_t$ as the explicit rope state, it is a little unclear to me what $x_t$ represents. Is this the Image? If yes, I would follow your previous notation for the image $I_t$. Using $x$ is confusing as it is typically used for system state. }
\jean{Reading on, you assume that $x$ is an observation of the state $s$. In that case I would go with $z$ for observation. }

\jean{Actually you now use the predicted state $\hat{s}_{t+1}$ as input to the sensor. So the below statement on independence is not quite true.}
We model the perception neural network as being part of the sensor, and the independent estimation of the rope state per image (i.e. without information from the previous frame) as the observation $x_t$ of state $s_t$. Thus the observation model $h$ is identity. \jean{It is a bit weird to formulate this as 'thinking of these matrices as diagonal'. Typically, $Q$ and $R$ are modelled as diagonal but the Jacobian $F$ and state covariance $\Sigma$ come out of the update equations or from the derivatives. Therefore they are not necessarily diagonal.} We also think of all matrices $F_t$, $\Sigma_t$ and $Q_t$ as diagonal matrices, although this assumption is not strictly true due to the constraint of constant rope length. Now we compare the scale of uncertainty from the process model and the observation model. The process model has a finite amount of uncertainty because the dynamics model is trained on simulated data, thus has a bias when used on real data. The observation model, i.e. the perception network, has negligible amount of noise compared to the process model when the rope segment is visible, since the network is well trained, and the estimated rope state can be refined with image loss to snap onto the rope pixels. However the perception model has much higher noise compared to the process model when the rope segment is occluded, since there is no information available in the image, and the network is only imagining based on its training data statistics. 

\jean{I'm a little uncertain if we really want to make this very strong claims about how this is all related to the EKF or even an approximation of it. The way you solve it is kind of funny when you relate it to an EKF. Not wrong, just funny. Because you actually do have a 'generative model' $h(\hat{s}_{t+1})$ that creates an expected observation. Just that this is now directly merged into the sensor (which is your network). Check out http://www.roboticsproceedings.org/rss10/p30.pdf on page 3. Maybe this is a way to put it?}

\jean{Also I'm wondering if you can simplify this discussion. Basically what you are saying is that given the prediction, you can easily do outlier rejection for parts of the rope and just not update it. This is almost as if you are cranking up the observation noise for parts of the rope. ANd this noise is estimated from the image given the prediction. To simplify things, I would just say that we assume some constant noise by the process model. The only thing that is varying is the observation noise.}
Based on the analysis, the EKF update rule is approximated to have a simple form. We will fully trust the perception network if the rope segment is visible, and fully trust the prediction from dynamics model if the rope segment is occluded. The latter half can also be achieved with our perception network after it is finetuned with our image loss. This is because the image loss does not have gradients on pixels that are occluded (Since we clip raw, negative gradients to zero), thus with a proper regularizer, the network will only update the provided coarse segment by linear interpolation if no rope pixel is in the ROI, essentially keeping the rope state intact. When doing tracking, the coarsest level prediction of $8$ segments are from the dynamics model prediction, thus the refined rope state will still be the same as predicted by the dynamics model, if the segments are occluded in the image.
}

\section{Data Collection}
\paragraph{Rendered images for perception model pretraining} We generated a dataset of 10000 rendered images, each containing a randomly generated b-spline curve with six control points. The b-spline curve is rendered with solid red color on white background. $65$ equidistant points are extracted from the spline as ground truth annotation of rope states. The length of the generated ropes range from $0.63$m to $1.25$m. 

\paragraph{Simulated dataset for training dynamics models} We generated a dataset of simulated rope manipulation sequences with ground truth rope states, actions and rendered images. The start state of each sequence is sampled from the dataset of 10000 b-splines used above. For each start state, we generate a $100$-step manipulation sequence. At each time step, a virtual robot grasps the rope at a random point and moves that point with a randomly generated displacement vector. The magnitude of random displacements is between $1$ and $3$cm. After the rope configuration is computed by the simulator, we render the rope from a top-down view using POV-Ray~\cite{pov-ray}. The images are only used for training the baseline video prediction model \cite{DVF_journal}. A total of $5800$ sequences are generated. We use $5112$ sequences for training the dynamics neural network and the rest for evaluations. 

\paragraph{Real dataset for perception model finetuning}
We also collected a dataset of real rope manipulation sequences in a similar manner. We used a Franka Panda robot arm \cite{Franka} with a parallel gripper to execute actions, and a Kinect camera to collect RGB images. The Kinect camera coordinate frame is calibrated to the robot base frame. Images collected from the Kinect camera are projected to top-down views using this transformation. No depth data is required for the calibration or projection. At the beginning of every recorded sequence, a human operator arranges the rope on the table. At each time step, the robot can choose to pick up the rope, move the gripper with rope being grasped, or release the rope and retract the arm from the table. For selecting the grasp location, a color filter is used on the Kinect image to segment the rope, and a random pixel in the rope segment is selected. Displacement vectors  for moving the gripper are selected at random as long as the gripper stays within the robot's workspace. Images are taken after each actions. For ease of transferring the pretrained network, the robot can move the gripper at most 5 times before releasing and retracting, so that we have un-occluded images of the rope spread through the recorded sequences. The real rope has length $1$m, and actions range between $1$ and $3$cm. We collected a total of $27$ sequences, with $26$ sequences totaling $5118$ images for training, transferring the pretrained perception model to the real environment using the self-supervising objective, and $1$ sequence with $201$ images for validation of the perception model. This dataset is also used to test the long-horizon prediction accuracy of our dynamics model, shown in Fig.6 (left) of the main paper. When using this data to evaluate the dynamics model, we use the state estimated by the finetuned perception network, further refined by directly optimizing the image loss w.r.t states, as the ground truth states.

\section{Additional Experiment Results}

\subsection{Tracking a rope making intersection}
\begin{figure}
    \centering
    \includegraphics[width=0.4\textwidth]{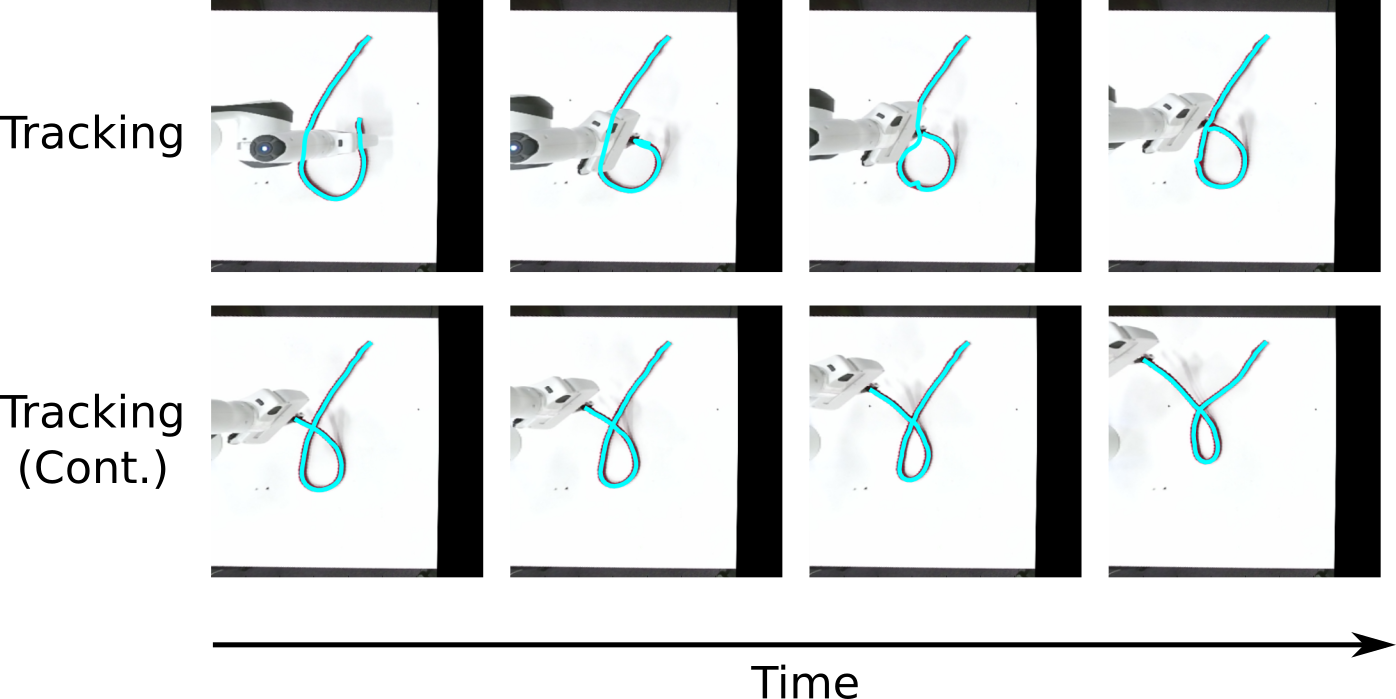}
    \caption{State estimation and tracking results on a manipulation sequence, where the robot actions are designed by the authors to make an intersection with the rope.}
    \label{fig:append_tracking}
\end{figure}

Visualization of the state estimation and tracking result on one manipulation sequence is shown in Fig.~\ref{fig:append_tracking}. The robot actions are designed by the authors to make an intersection (topological change). The tracking method is described in Sec. IV(D) of the paper. This result demonstrates that our perception method can also be used in more complex rope manipulation tasks, such as knotting.

\subsection{Network finetuning with curriculum learning}
\begin{figure}
    \centering
    \includegraphics[width=0.35\textwidth]{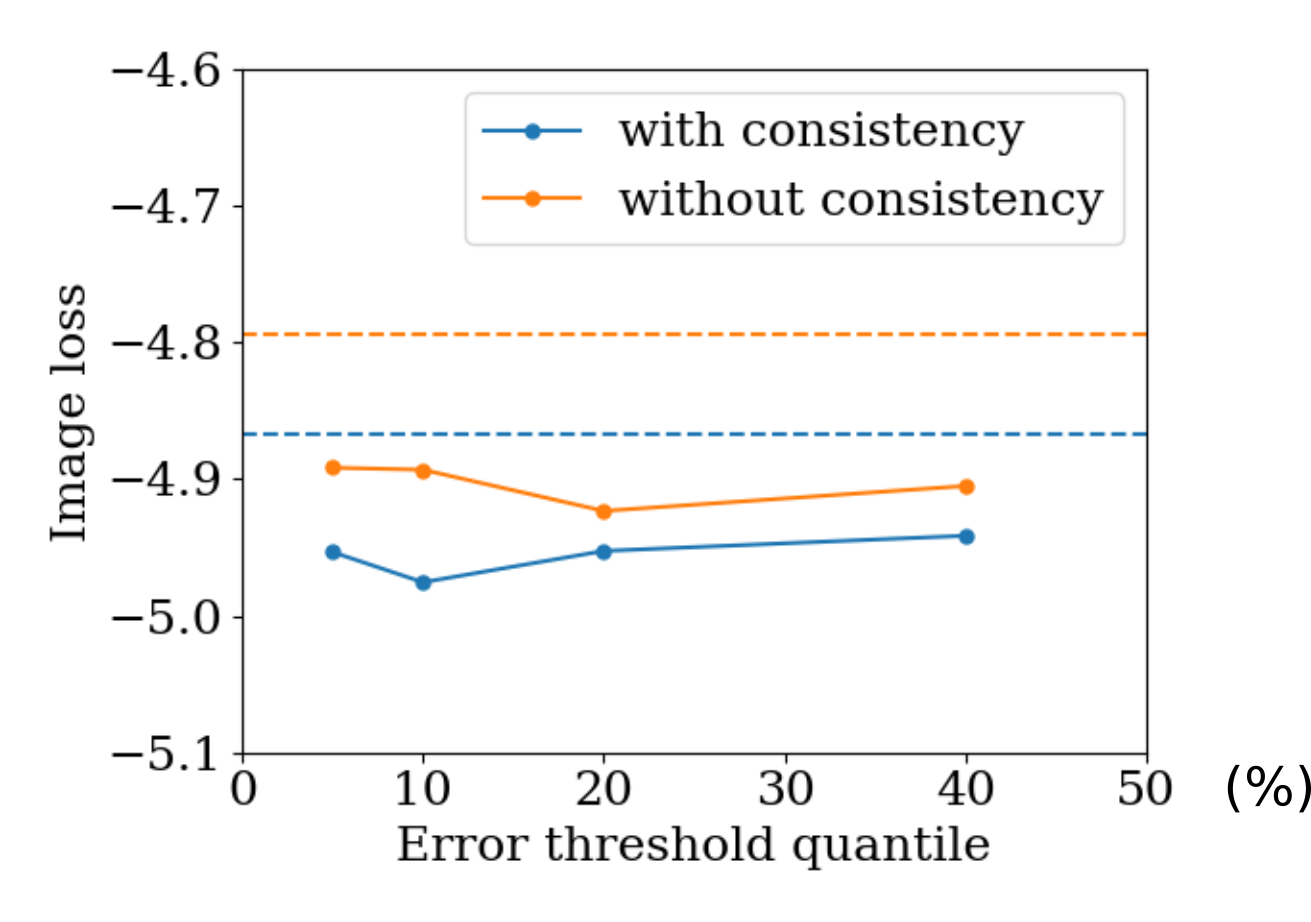}
    \caption{Final image losses on the training set for finetuning with different error thresholds in automatic curriculum learning. The dashed straight lines are the final image losses when trained without using curriculum learning.}
    \label{fig:append_curriculum_compact}
\end{figure}

We provide more detailed experiment results on the effects of choosing different error thresholds for the automatic curriculum learning. We choose error thresholds that correspond to the top 5\%, 10\%, 20\%, and 40\% quantile image loss among the training dataset before finetuning. The final average image losses after convergence are shown in Fig.~\ref{fig:append_curriculum_compact}. The marked dots represent training results for different error thresholds, and the dashed lines provide baseline results where curriculum learning is not used. The figure shows it is best to use an error threshold between the 10\% to 20\% quantile of the training dataset, however the differences among different error thresholds are small compared to the improvement upon the baselines, thus demonstrating that using automatic curriculum learning is effective for a wide range of error thresholds used.

\begin{figure}
    \centering
    \includegraphics[width=0.35\textwidth]{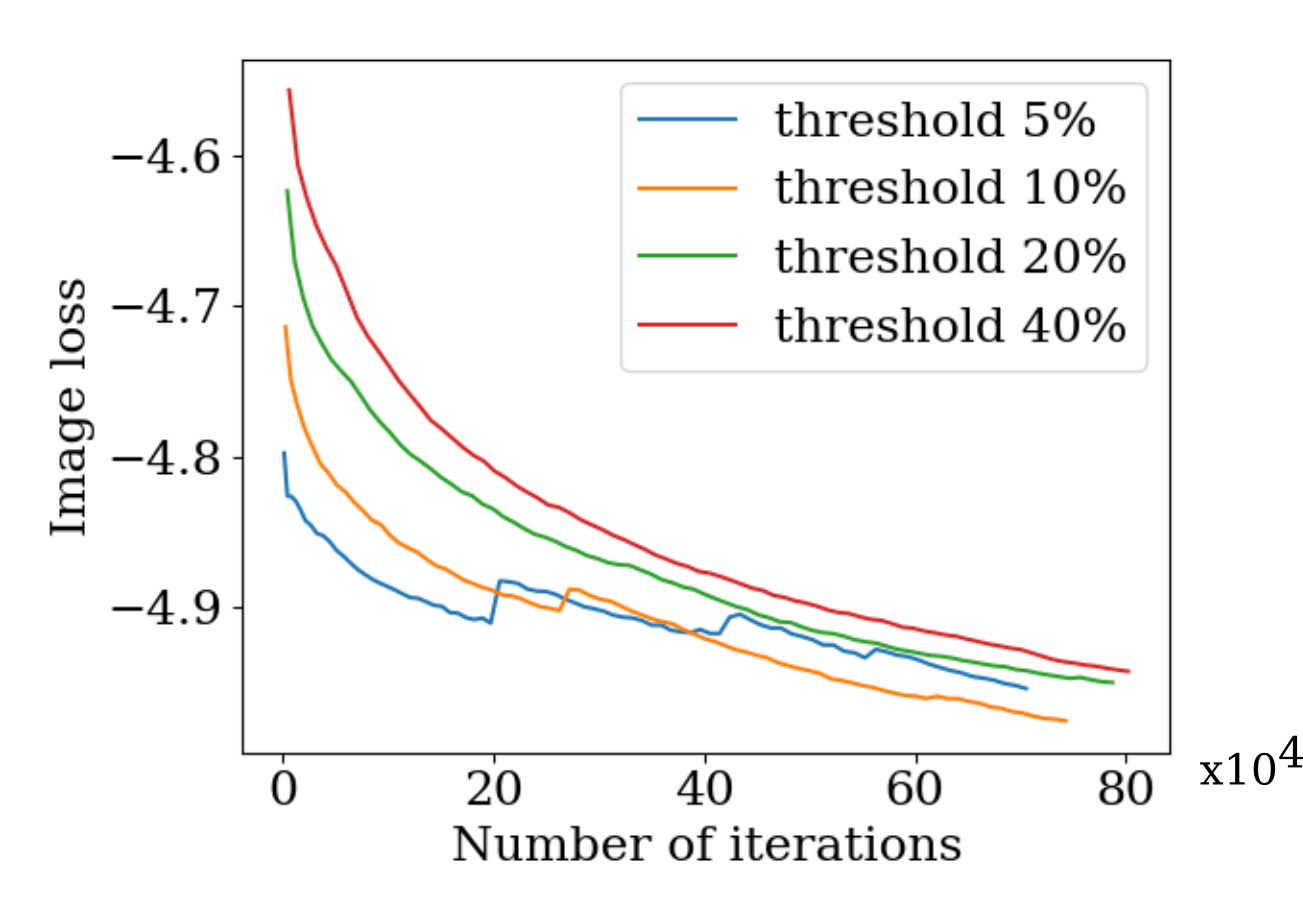}
    \caption{The training loss curves for finetuning the network with both automatic curriculum learning and temporal consistency. The reported image losses are averaged over the effective training set, i.e. samples whose loss is below the error threshold. Stairs in the training loss curves correspond to when the error threshold is automatically adjusted to include more training samples.}
    \label{fig:append_curriculum_train}
\end{figure}

We also include the training curves when training with both automatic curriculum learning and temporal consistency, where the error threshold for curriculum learning varies. The training curves are shown in Fig.~\ref{fig:append_curriculum_train}. The reported image losses are averaged over effective training samples, i.e. samples whose losses are below the error threshold. Therefore, the initial training losses for stricter error thresholds are lower. When training with a stricter error threshold, there are stairs in the training curve, corresponding to when the numbers of effective training samples have stopped increasing, and the error thresholds are automatically updated to include more training samples.

\subsection{Data efficiency of dynamics models}
\begin{figure}
    \centering
    \includegraphics[width=0.35\textwidth]{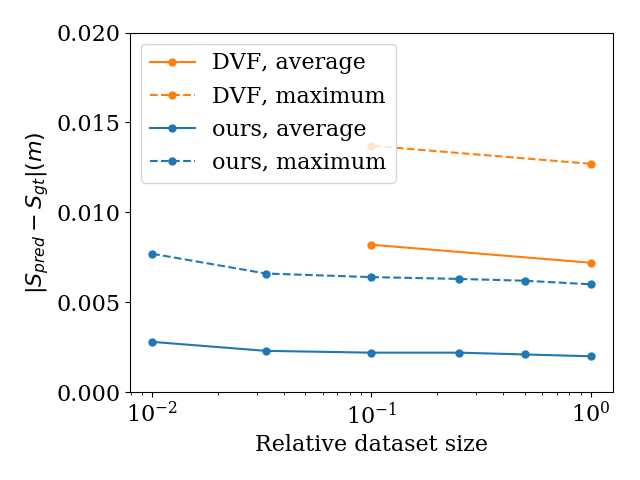}
    \caption{Average and maximum deviation from predicted states to the ground truth on the evaluation set, when the training data size varies for both our LSTM model and the baseline model. The data sizes are measured as relative sizes compared to the largest dataset we generated, which contains 0.5M simulated actions.}
    \label{fig:append_data_efficiency}
\end{figure}

We provide more experiment results to demonstrate that our dynamics model using bi-directional LSTM has much higher data efficiency compared to the baseline model~\cite{DVF}. In Fig.~\ref{fig:append_data_efficiency}, we plot the average and maximum deviation from the predicted states to the ground truth states on the evaluation set, when the training data size varies. The training data sizes are reported as relative sizes compared to the biggest dataset we generated, which contains 0.5M simulated actions, as described in the previous section. The results show that our dynamics model is able to train well with only 3\% of the total training data, only showing overfitting and worsened results when training data size reduces to 1\%. On the other side, the baseline model would need larger datasets to continue to improve its performance, which we do not have enough resources to generate.

\subsection{Simulated manipulation experiments visualization}
\begin{figure}
    \centering
    \includegraphics[width=0.4\textwidth]{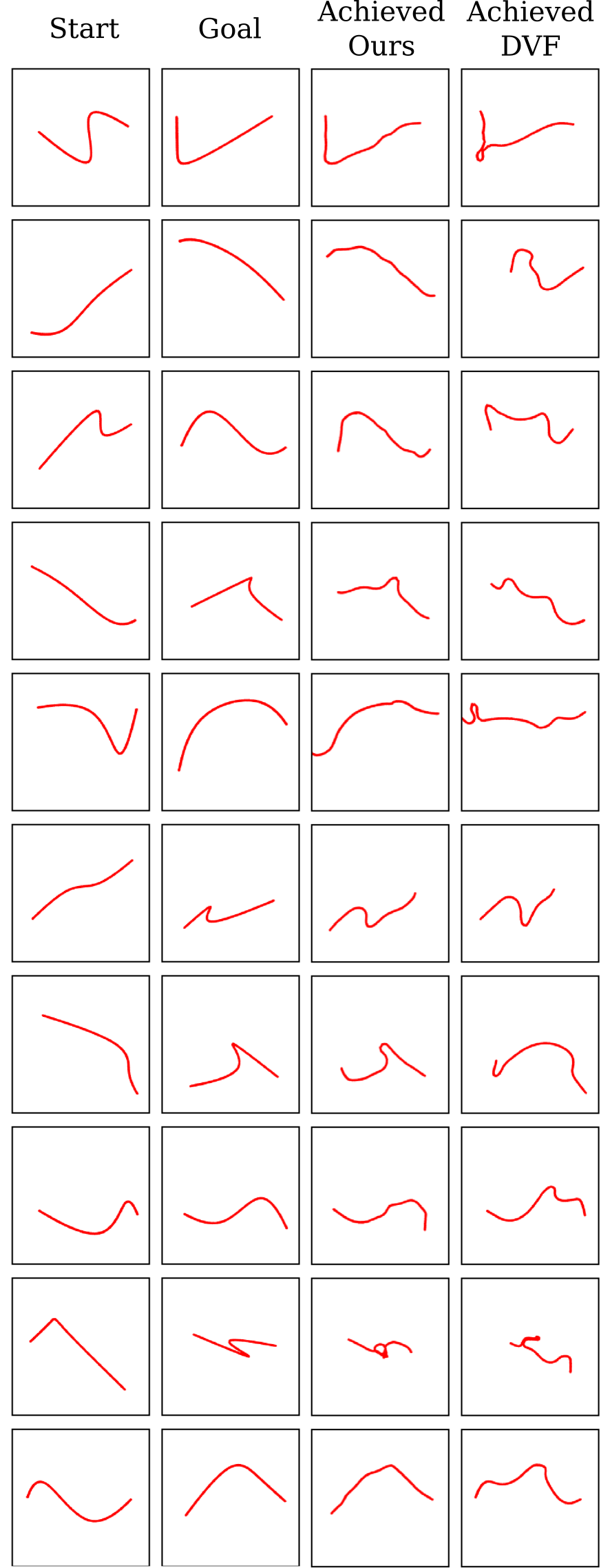}
    \caption{10 example manipulation experiments in simulation. From left to right: the start state, the specified goal state, state achieved with our method at $t=100$, and state achieved with the baseline method(DVF)\cite{DVF_journal} at $t=100$.}
    \label{fig:append_combine}
\end{figure}

We include visualizations of the simulated manipulation experiments, described in Sec. IV(D) of the main paper. The start state, specified goal state, and states achieved by each method after $100$ actions are shown in Fig~\ref{fig:append_combine}. Our method achieves states much closer to the specified goals.

\bibliographystyle{IEEEtran}
\bibliography{IEEEabrv,ref}

%% file: introduction.tex
\begin{figure}[h]
    \centering
    \vspace*{6pt}
    \includegraphics[width=0.45\textwidth]{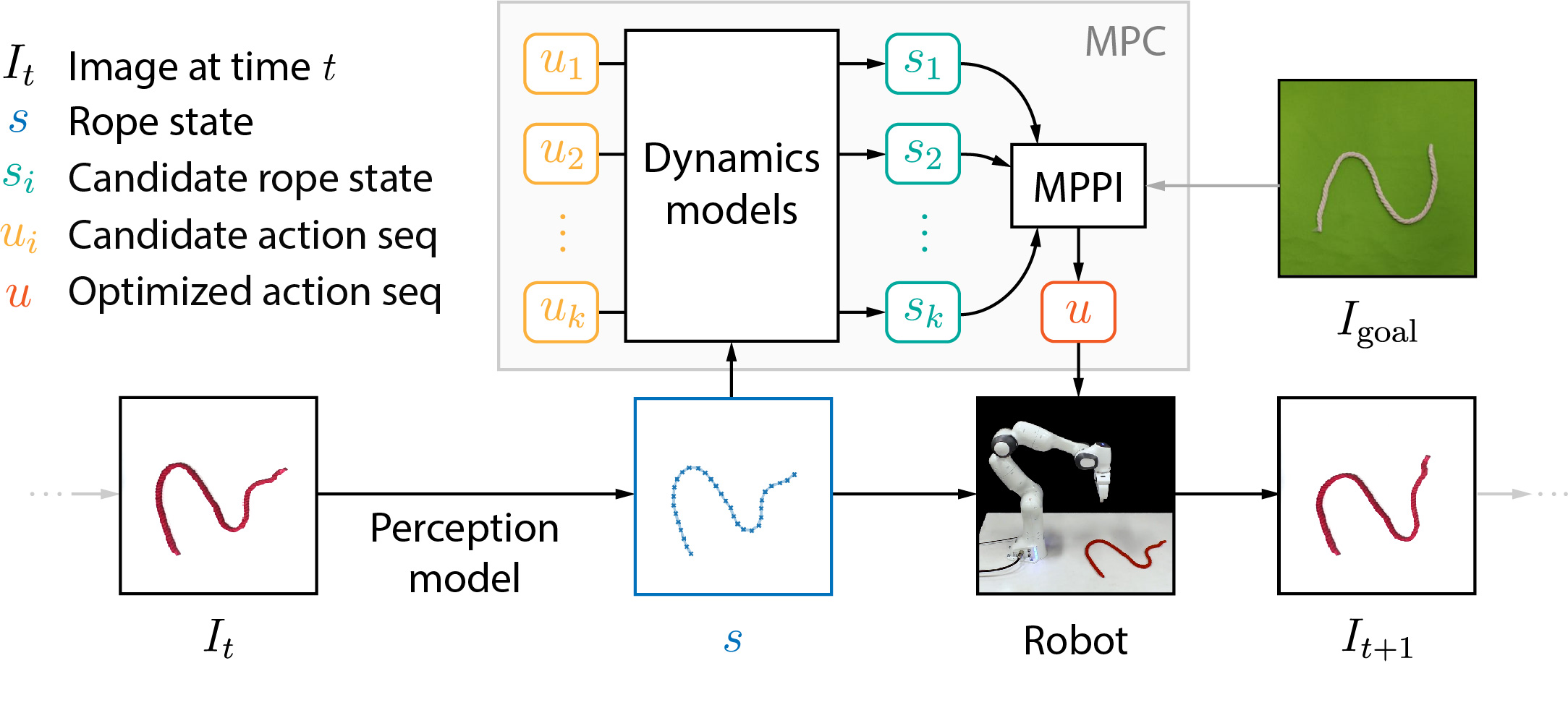}
    \caption{Overview of our rope manipulation system. Given an image, our perception model estimates the explicit rope state. The state is refined by minimizing our proposed self-supervising objective w.r.t. the rope state. A dynamic model in rope state space predicts future states given the current state and hypothetical action sequences. MPPI is used to optimize action sequences according to the distance from predicted states to the goal state, also estimated from an image. The robot executes the first few actions and obtains a new observation. The state estimated from that image provides the input to MPPI to replan. This repeats until the goal state is reached.}
    \label{fig:overview}
\end{figure}

Manipulating deformable objects is an important but challenging task in robotics. It has a wide range of applications in manufacturing, domestic services and health care such as robotic surgery, assistive dressing, textile manufacturing or folding clothes \cite{Suture, Dressing, Folding}. Unlike rigid objects, deformable objects have a high-dimensional state space and their dynamics is complex and nonlinear. This makes state estimation challenging and forward prediction expensive.

We propose a vision-based system that allows a robot to autonomously manipulate a linear deformable object to match a visually provided goal state. Previous learning-based approaches towards this problem \cite{ZSVI, DVF_journal} learn models in image space or a latent space, and do not incorporate any physics prior into the learning process. While conceptually these methods could be applied to other object classes, they suffer from low data efficiency and difficulty in generalization~\cite{kloss_icra2018}. 
We take a different approach to this problem, based on an explicit state-space representation of the physical system. This choice has several advantages. First, it allows us to incorporate physics priors about the behaviour of a deformable object when it is manipulated, e.g. by reflecting a mass-spring system in the network structure. As we show in our experiments, such dynamics models produce more realistic predictions of the object's behaviour over a longer horizon than dynamics models learned directly from pixels. Second, explicit states are invariant to the appearance of the object and its environment. Therefore, dynamics learned in one setting can be directly used in other visually different settings. It also allows us to specify goal shapes with one rope that is then achieved with a rope of different length, thickness, and/or appearance. 
It is not obvious how to achieve this invariance with a method operating in a learned latent space or in pixel space. Finally, an explicit state-space representation more readily lends itself to manipulation planning and control especially when optimizing a sequence of actions. It is straight-forward to construct intuitive and informative losses for the optimization, as well as heuristics of promising action sequences to initialize the optimization. 

The main challenge then becomes to estimate this explicit state from raw images. This task has previously been tackled by hand-engineered image processing algorithms, e.g. in \cite{DeformPercept1}. Recently, Pumarola et al.~\cite{DeformPercept2} demonstrated explicit state estimation for deformable surfaces in simulation, where ground truth annotations are easily accessible. Such annotations are expensive to obtain for real images. We overcome this problem by proposing a self-supervising learning objective that enable continuous, self-supervised training of deformable object state estimation on real images once the model is initialized with a small set of synthetic images.

We demonstrate the effectiveness of our method on the task of rope manipulation. We embed the learned perception model for explicit state estimation into a full system that includes a dynamic model in state space with physics priors, and {\em Model Predictive Control\/} (MPC), shown in Fig.~\ref{fig:overview}. We quantitatively show that the proposed method is significantly more efficient in manipulating ropes to match specified goal configurations compared to previous methods that learn in pixel space.


Summarizing, the contributions of this work are:\\
(i) We propose a novel self-supervising learning objective
that enables training state estimation on real data, without requiring expensive ground truth annotations.\\
(ii) We propose coarse-to-fine state estimation using hierarchical Spatial Transformer Networks (STN), which shows improved generalization compared to direct state estimation.\\
(iii) We propose a novel dynamics model in state space that enforces physics priors for linear deformable objects. Our dynamics model has comparable performance to a physics simulation engine, while being significantly faster. Our model reduces prediction error by 68\% while only using 3\% of total training data, compared to a baseline dynamics model in pixel space~\cite{DVF_journal}.\\
(iv) We demonstrate rope manipulation in both simulated and real environments. Quantitative comparisons in simulation show significantly more efficient manipulation of our method compared to a baseline.

%% file: related.tex
\paragraph{Self-supervised State Estimation}


While a lot of previous works learn dynamics models in image space or latent space using self-supervision, only a few have looked at self-supervised learning of explicit state estimation, such as object pose estimation. Wu et al.~\cite{Wu_NIPS2017} used a differentiable renderer to convert predicted rigid object poses back to images, and compare with ground truth observations. Ehrhardt et al.~\cite{Ehrhardt2018} proposed two regularizing losses to enforce object trajectory continuity and spatial equivariance. Byravan et al. \cite{SE3PoseNets} train networks that learn robot's latent link segmentation and pose space dynamics from point cloud time series, using a reconstruction loss for self-supervision.
Our image-space loss is inspired by this line of works, but addresses much higher-dimensional linear deformable objects like ropes. We are able to train the perception network on real data with only self-supervision, provided that it has been warm started with a small set of rendered images.

\paragraph{Deformable Object Tracking}
Several previous works have studied tracking of deformable objects given segmented point clouds \cite{Tracking_ICRA13, Tracking_IROS15, Tracking_IROS17}. On the high level, these tracking methods iteratively deform a pre-defined mesh model to fit the segmented point cloud, and use physical simulation to regularize the mesh deformation to have low energy. There are two major limitations in the previous tracking methods: (1) the mesh configuration needs to be initialized manually or using algorithms engineered for specific cases. (2) Rope segmentation is required as input, usually achieved by color/depth filtering, assuming the foreground and/or background has a known solid color. Our state estimation method eliminates the need for manual initialization by introducing a perception neural network. We also alleviate the assumption of known foreground/background colors, only requiring that the foreground and background have good color contrast, to achieve self-supervision on real images. Our state estimation method is generalizable to a range of visually different objects and backgrounds.

\paragraph{Deformable Object Manipulation} 
There are two main branches of work in the area of deformable object manipulation other than rope manipulation. One branch of work focuses on clothing items~\cite{Malassiotis2016clothing,Li2016clothing}. Researchers designed robot action strategies specific for each of the task stages, and formulated the object states as key point positions and object contours, to drastically simplify the complex problem. Our work tackles the simpler problem of linear object manipulation, proposing a much more generalizable solution, with task-independent state space and action space formulations. The other branch of work focuses on flexible beams or cables~\cite{Moll2006path,Roussel2015rod}. These works assume that the elastic forces are much stronger than gravity or other external forces, such as friction, and the shape of the beam/cable can be fully determined by moving or twisting the two end-points. This model does not apply to soft ropes. The shape of ropes heavily depend on the manipulation history due to friction with the supporting plane. Furthermore, ropes are not fully controllable by just grasping the two end-points.

\paragraph{Rope Manipulation}
Specific to rope manipulation, Yamakawa et al.~\cite{DynamicKnot} have demonstrated high-speed knotting, which depends on an accurate dynamic model of the robot fingers and the rope. Lee et al.~\cite{TPS} and Tang et al.~\cite{TPS-T} use spatial warping to transfer demonstrated manipulation skills to new but similar initial conditions. More related to our work, Li et al.~\cite{PropNet} and Battaglia et al.~\cite{InteractionNet} model ropes as mass-spring systems, and use graph networks to learn rope dynamics. However, they assume that the rope's physical state is fully observable.
Ebert et al.~\cite{DVF_journal} learn a video prediction model, without any physical concept of objects or dynamics. However, a series of efforts \cite{Chelsea17, Chelsea18} has been made to find informative losses on images, which are required for long-horizon planning in the model predictive control framework. 
Wang et al.~\cite{Abbeel_RSS19} embed images into a latent space associated with an action-agnostic transition model, plan state trajectories and servo the trajectory with an additional learned inverse model. While they achieve satisfactory result manipulating one particular rope, visually different ropes are not guaranteed to share the embedding space and transition model, making generalization difficult. Different from previous works, we directly estimate the ropes' explicit states from images. This space lends itself more readily to efficient learning, flexible goal specification and manipulation planning than pixel space or latent spaces.

%% file: method.tex

A flow chart of the full system at test time is shown in Fig.~\ref{fig:overview}. We use a robot arm with gripper to manipulate a rope on the table. The task is to move the rope to match a desired goal state, specified by an image. At each time step, a {\em Convolutional Neural Network\/} (CNN) estimates the explicit rope state. 
The structure of this network is described in Sec. \ref{sec:perception}. The network is first trained with rendered images, then finetuned on real images with our proposed self-supervising learning objective described in Sec. \ref{sec:imageloss}. With the estimated states, we use {\em Model Predictive Path Integral Control\/} (MPPI) \cite{MPPI} in combination with a dynamics model to optimize a sequence of actions.
Our proposed dynamic model is described in Sec. \ref{sec:dynamics}. Details on the modified MPPI algorithm are described in Sec. \ref{sec:mpc}.

\subsection{Coarse-to-fine State estimation}
\label{sec:perception}
\begin{figure}[t]
\centering
\vspace*{0.08in}
\includegraphics[width=0.45\textwidth]{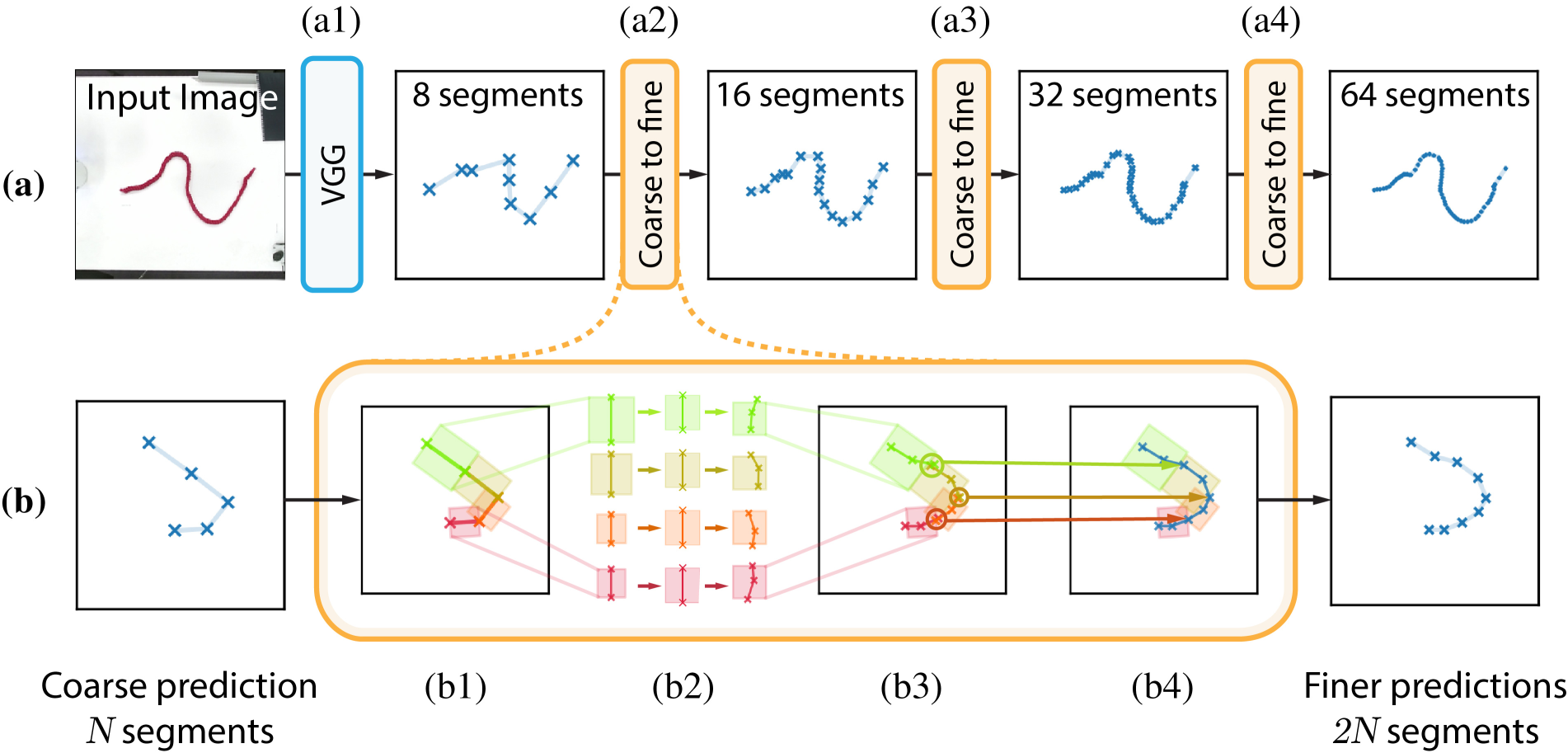}
\caption{Coarse-to-fine estimation of rope state. (a1) Given an input image, the neural network first estimates 8 straight segments. (a2-a4) The segment estimations are hierarchically refined using STNs. (b1) Square  boxes are defined by the previously estimated segments. (b2) The boxes are used to extract regions from the VGG feature maps, and fed into a multi-layer perceptron to estimate the left endpoint, middle point, and right endpoint in each square region. The estimated points are concatenated(b3) and endpoints from neighboring regions are averaged(b4) to obtain a higher
resolution estimate, based on twice the amount of segments that more closely model the shape of the rope.}
\label{fig:state_estimation}
\end{figure}

We formulate the problem of rope state estimation from an RGB image as estimating the positions of an ordered sequence of points on the rope. The rope state estimation problem has a divide-and-conquer structure, i.e., estimating the state of a segment of the rope is the same problem as estimating the state of the entire rope. To exploit this structure, we use STNs \cite{STN} to estimate the rope state in a coarse-to-fine manner as visualized in  Fig.~\ref{fig:state_estimation} (a). A VGG network \cite{vgg} first estimates 8 straight segments that roughly approximate the shape of the rope. Using STNs, 8 square regions are extracted, one per segment. Within each extracted region, the network updates the position of the two end-points and estimates the position of the middle point on the rope segment. The outputs are converted back to the original image coordinates, and endpoints from neighboring regions are averaged, so that the entire rope is now represented by 16 straight segments (see Fig.~\ref{fig:state_estimation} (b)).  New regions are extracted for each of the new segments, with higher spatial resolution. The process repeats until the rope is represented at sufficient resolution, in our case with 64 segments. The detailed network structure and parameters are described in Appendix (I). Training and testing data for this and following networks are described in Appendix (V). Code will be made available upon publication.

\subsection{Self-supervising learning objective}
\label{sec:imageloss}
\begin{figure}
    \centering
    \vspace*{0.08in}
    \includegraphics[width=0.35\textwidth]{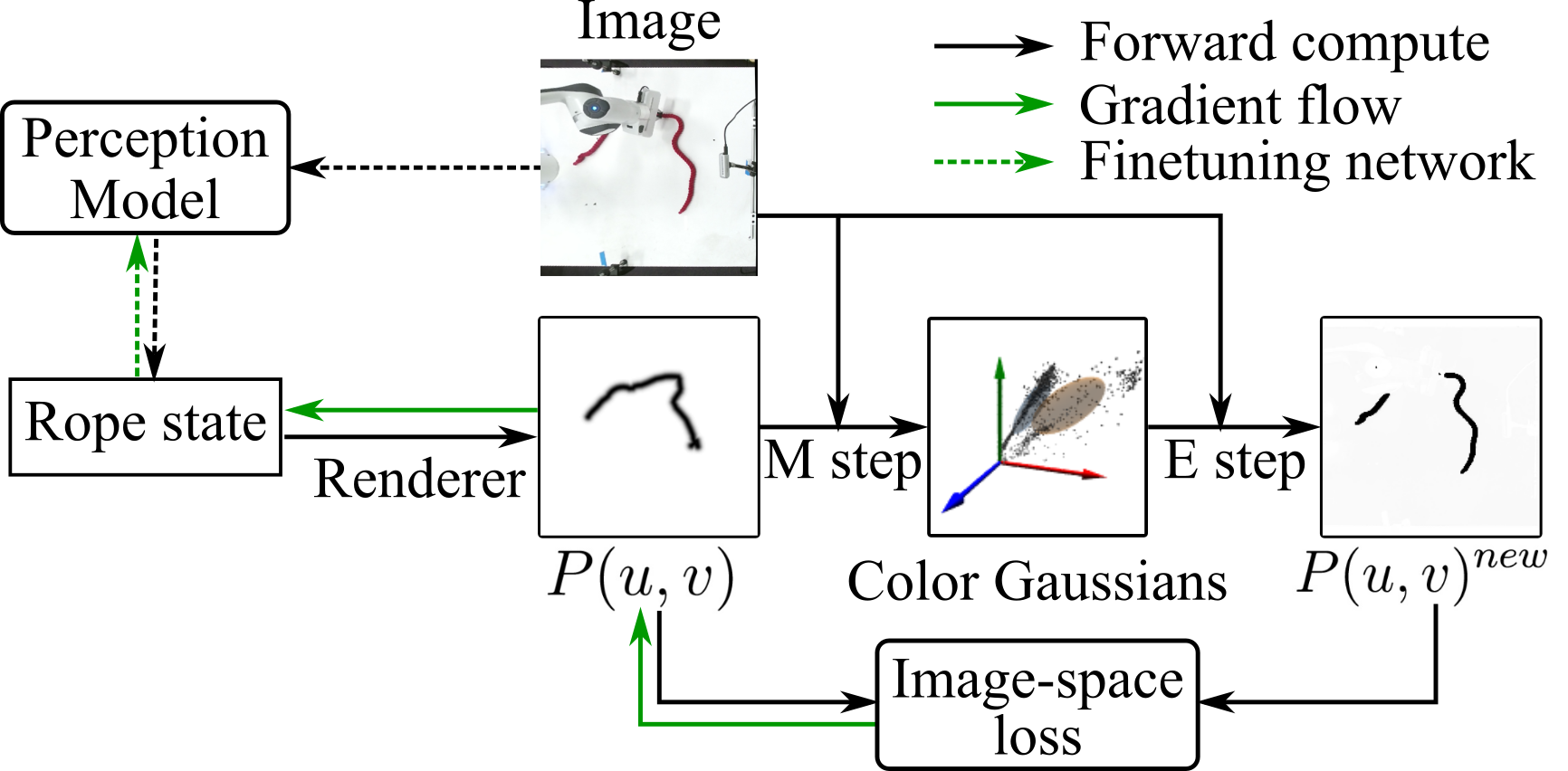}
    \caption{ Illustration of our proposed self-supervising learning objective. From an input image and initial rope state estimate, a differentiable renderer computes the membership weights $P(u,v)$. Then the M step produces GMM parameters, and the E step produces new membership weights $P(u,v)^{new}$. Given this, the image loss between $P(u,v)$ and $P(u,v)^{new}$ (Eq.~\ref{eq:image_loss}) is minimized to either update the rope state, or to update the weights of the perception network (dashed green line).}
    \label{fig:imageloss_vis}
\end{figure}

We train the neural network model with rendered spline curves as ropes, where ground truth rope states are easily available. However, real images look different from rendered images in many aspects, e.g. different lighting, distractor objects, occlusions through robot arms or, a different rope shape distribution. To close the reality gap without requiring ground truth annotations of rope states in real images, we propose a novel learning objective, consisting of a differentiable renderer and an image space loss to achieve self-supervision on real images (see Fig.~\ref{fig:imageloss_vis}).

Our method makes the assumption that the object has a good color contrast with the background, which is often the case for ropes or other linear deformable objects. Consider the simplifying case where the rope and the background each have a solid color. If we think of each pixel as a point in RGB space, all the pixels should form two clusters, one around the rope color and one around the background color. If we model the distribution in RGB space as a mixture of two Gaussians, and assign each pixel to the more probable Gaussian, the assignment variables will give us the segmentation mask of the rope versus background. When we estimate the rope configuration using the perception network, this estimate should agree with the color-based segmentation.

Clustering in RGB space can be achieved with the {\em Expectation-Maximization\/} (EM) algorithm for {\em Gaussian Mixture Models\/} (GMM). Given an initial estimate of GMM parameters $\Theta$, i.e., the component weights $\alpha_k$, means $\mu_k$, and covariance matrices $\Sigma_k$, $1 \leq k \leq K$, the EM algorithm iterates between the E step, which updates the membership weights $w_{ik}$ of data point $x_i$ to cluster $k$, and the M step, which updates $\Theta$. 
In the E step, membership weights are updated as:

\vspace{-3pt}
\begin{equation}
\label{eq:e-step}
    w_{ik} = \frac{\alpha_k p_k(x_i | \mu_k, \Sigma_k)}{\sum_{m=1}^K \alpha_m p_m(x_i | \mu_m, \Sigma_m)}
\end{equation}

where $p_k$ and $p_m$ are multivariate Gaussian densities. In the M step, GMM parameters are updated as:
\begin{equation}
\label{eq:m-step}
\small
\begin{aligned}
    \alpha_k^{new} &= \sum_{i=1}^N w_{ik} / N, \\
    \mu_k^{new} &= \sum_{i=1}^N w_{ik}x_i / \sum_{i=1}^N w_{ik},  \\
    \Sigma_k^{new} &= \sum_{i=1}^N w_{ik}(x_i-\mu_k^{new})(x_i-\mu_k^{new})^T / \sum_{i=1}^N w_{ik}.
\end{aligned}
\end{equation}

For rope state estimation, we model the distribution in RGB space with two Gaussians, thus $K=2$. For each pixel with coordinate $(u,v)$, let $P(u,v)$ be its membership weight to the rope RGB cluster parameterized by $\mu_1$ and $\Sigma_1$, i.e., $w_{i1}$. Then $1-P(u,v)$ is the membership weight of pixel $(u,v)$ to the background RGB cluster parameterized by $\mu_2$ and $\Sigma_2$, i.e., $w_{i2}$. $x_i$ refers to the RGB value of pixel $(u,v)$. While the M step is straightforward to apply given $P(u,v)$, the per-pixel membership weights $P(u,v)$ should be expressed in terms of the estimated rope state, instead of i.i.d. per pixel. Thus, the E step does not apply as is. 

We propose a differentiable renderer that links $P(u,v)$ to the rope state. The rope state is a sequence of $64$ segments. We individually render each segment with end-points $p_j$, $p_{j+1}$  to get $P_j(u,v)$, and take the pixel-wise maximum, $P(u,v)=\max_jP_j(u,v)$. Rendering of one segment is defined as 
\begin{equation}
\label{eq:render}
    P_j(u,v) = \exp{(-d_j(u,v)^2/\sigma^2)}
\end{equation}
where $d_j(u,v)$ is the distance of pixel $(u,v)$ to its closest point on segment $j$.  $\sigma$ is a learnable parameter that controls the width of the rendered segments.

Given the initial state estimate from the neural network, we can compute $P(u,v)$ based on Eq.~\ref{eq:render}. The M step can be applied easily to compute the parameters of the Gaussian clusters in RGB space. We also follow the E step in Eq.~\ref{eq:e-step} to calculate $P(u,v)^{new}$. Then, instead of directly using $P(u,v)^{new}$ for the next M step, we refine the estimated rope state by minimizing the distance between $P(u,v)$ and $P(u,v)^{new}$, defined as
\begin{equation}
\label{eq:image_loss}
\footnotesize{
    \sum_{(u,v)}-\log\left[P(u,v)P(u,v)^{new}+(1-P(u,v))(1-P(u,v)^{new})\right].}
\end{equation}
Thus, we adapt the EM algorithm for GMMs to minimize the above loss. Note that, the GMM parameters are only transient values estimated for each individual image. They are not memorized as parameters. Thus our method does not make any assumption about the distribution of rope colors or background colors, which was used in many previous works~\cite{Tracking_ICRA13, Tracking_IROS17, Malassiotis2016clothing, Li2016clothing}. Instead, we make the much weaker assumption that the rope has good color contrast with the nearby background.

To model occlusions, e.g. from the robot arm, we clip the gradient of this loss on each pixel $P(u,v)$ to be non-negative. In this way, we do not penalize pixels that belong to a rope segment according to the estimated rope state, but whose color belongs to the background color cluster, because that rope segment could be occluded. 

The proposed loss can be used for either finetuning the perception network or for refining the rope state estimate at test time, without updating the network weights.

\begin{figure}
    \centering
    \includegraphics[width=0.4\textwidth]{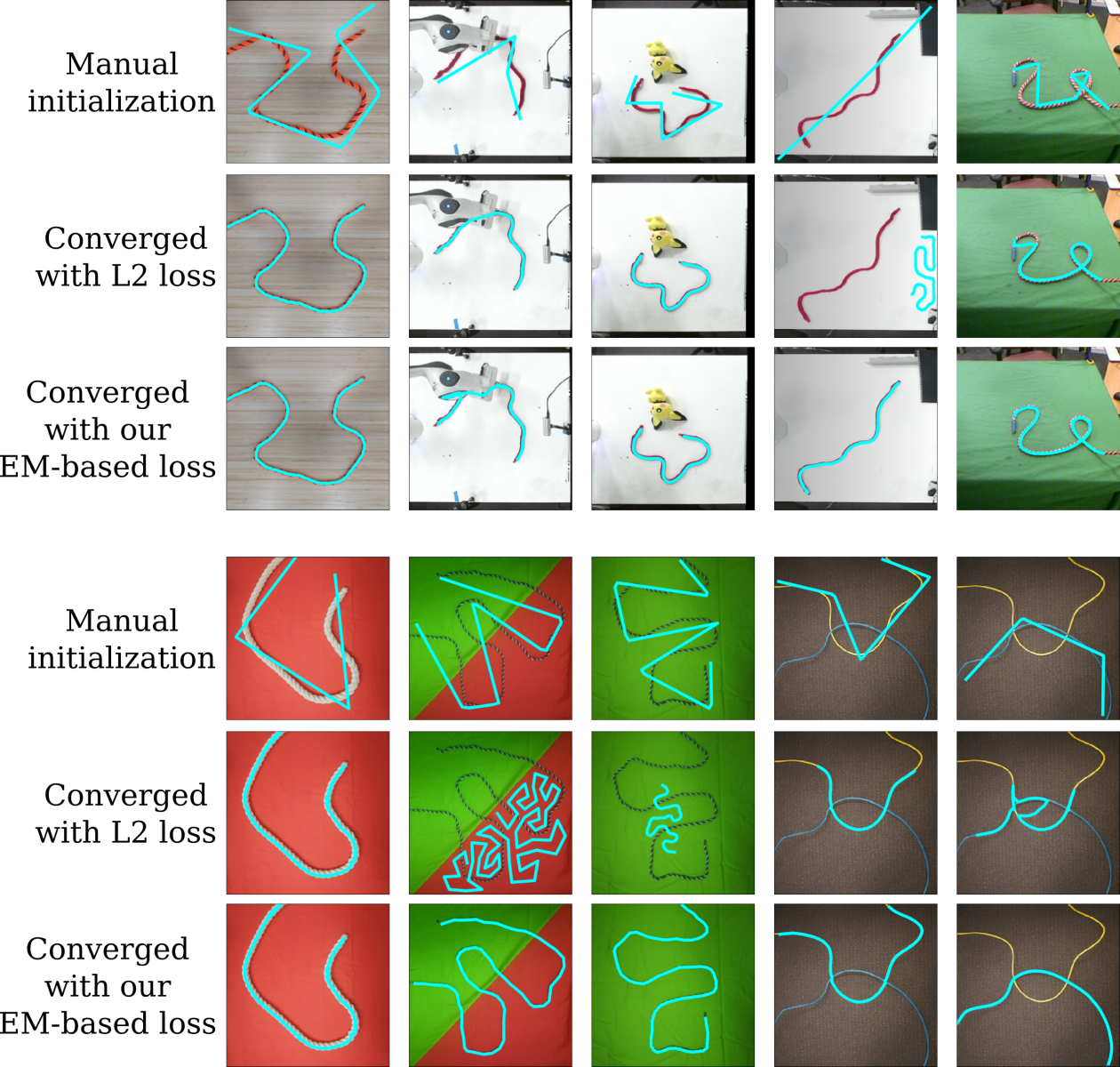}
    \caption{Generalizability of self-supervising objectives in refining initial rope state estimations. Rope states are overlaid on the input images as blue lines. Our proposed objective is robust to lighting variation, robot occlusion, distracting objects, ropes/backgrounds with more than one color, and the presence of multiple ropes. Using the L2 loss failed to converge to the desired result on some real images.}
    \label{fig:compare_image_loss}
\end{figure}

\paragraph{Network finetuning with an automatic curriculum and  temporal consistency}
While the self-supervising objective is generalizable across different visual appearances, it is not free of local minima. When the estimate from the perception network is not good enough, gradients of the proposed objective could lead the rope state into undesired local minima. When finetuning the perception network on real images, we want to prevent such undesired gradients from negatively affecting the network weights. We use an automatic curriculum based on the current loss of each training example. 
Training examples whose current loss is above a per-determined threshold are ignored during gradient updates. Since the selected examples are already very close to the true rope state thus having low loss, it is very unlikely their gradients will lead to wrong local minima. As the network learns, examples that originally have higher losses will improve, and their probability of falling into undesired local minima decreases. These examples will be included in the effective training set at a later point when their losses drop below the threshold. 

In addition to using a curriculum, we also exploit temporal consistency in the recorded sequences to help the learning converge faster and better. If one frame has a loss below the threshold while its neighboring frame has a loss above the threshold, we take the predicted rope state from the better frame to guide the prediction on the worse frame. 
Exploiting temporal consistency in self-supervised training greatly improves the result, as shown in Fig \ref{fig:finetuning_image_loss}. For more details about the finetuning algorithm, refer to Appendix (II).

\comment{
\begin{figure}
    \centering
    \includegraphics[width=0.32\textwidth]{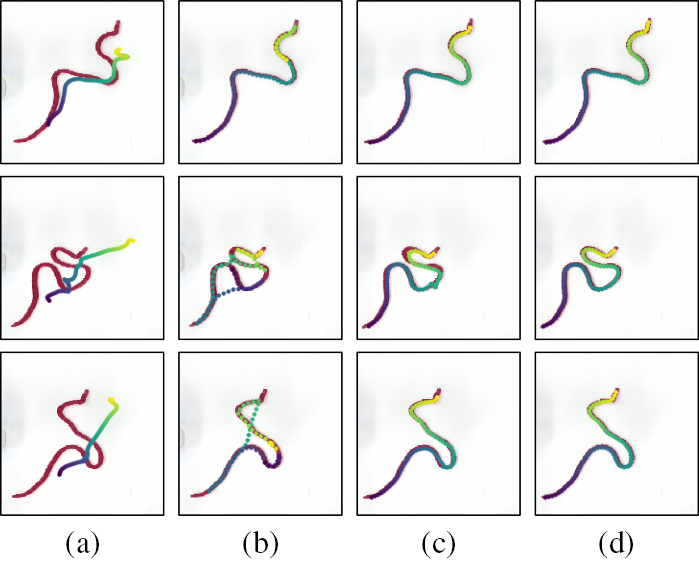}
    \caption{The benefit of using automatic curriculum in finetuning the perception network. (a) Hand-picked bad state estimates from network trained only on simulated data. (b) Refining (a) using image loss leads to undesired local minima. (c) State estimate from network finetuned with image loss and automatic curriculum. (d) Refining (c) using image loss no longer suffers from undesirable local minima.}
    \label{fig:local_minima}
\end{figure}
}

\paragraph{Generalization to more complex visual appearances}
Although we derived the objective for the simple case of rope and background each having a solid color, we note that this objective is also applicable if the rope or the background is textured with several different colors, or when there are distractor objects or occlusion. We demonstrate a few examples in Fig.~\ref{fig:compare_image_loss}. 
To demonstrate the effectiveness of our objective independent of other components, we manually initialized the rope state estimation by clicking a few points on the images. During manipulation experiments, such initial estimates are provided by the perception network. The refined rope state estimate after convergence is shown in Fig.~\ref{fig:compare_image_loss} ($3^{rd}$ and $6^{th}$ row). We compare to the method generalized from \cite{Wu_NIPS2017}, where the rendered grey scale image $P(u,v)$ is colored with the mean color of each cluster, and the L2 loss with the input image is used (Fig.~\ref{fig:compare_image_loss} ($2^{nd}$ and $5^{th}$ row)). Our proposed objective is robust to lighting variations, bi-colored ropes, bi-colored/textured backgrounds, distractor objects, multiple ropes, or occlusions. The baseline method does not always converge to the desired solution. The superior robustness of our method compared to using the L2 loss can be attributed to the different assumptions used by GMM clustering and K-means clustering. Using the same notation as above, the L2 loss can be written as:

\vspace{-5pt}
\footnotesize
\begin{eqnarray*}
& &\sum_i \|x_i - w_{i1}\mu_1 - w_{i2}\mu_2\|^2 \\
&=& \sum_i w_{i1}\|x_i - \mu_1\|^2 + w_{i2}\|x_i - \mu_2\|^2 + \sum_i w_{i1}w_{i2}\mu_1^{T}\mu_2.\\
\end{eqnarray*}

\vspace{-5pt}
\normalsize
Since $w_{i1}, w_{i2}$ are computed from Eq.\ref{eq:render}, $\sum_i w_{i1}w_{i2}$ only depends on the hyper parameter $\sigma$ and the total length of the rope when approximated to the first order. Thus we do not consider the effects of the last term. The first two terms correspond to K-means clustering in the RGB space. K-means clustering has the following assumptions: (1) The variance of each cluster is roughly equal. (2) The number of data points in each cluster is roughly equal. (3) The distribution of each cluster is roughly isotropic (spherical). All of these assumptions can be broken in real images, e.g. when the rope or background has multiple colors, or when the variance of brightness of pixels is much larger than the variance of hue, due to lighting and shadows.
On the other hand, our proposed loss is using GMM to cluster the RGB space, thus is more robust on real images.

\subsection{A dynamics model with a physics prior}
\label{sec:dynamics}

After rope states are estimated by the perception network and further refined with the proposed objective, a dynamics model is needed to predict future rope states given hypothetical actions, so that we can plan action sequences towards the goal. 

Although physics-based simulators for deformable objects are available~\cite{physbam}, we train a neural network dynamics model to speed up inference and support parallel processing.
To encode the physics prior of linear objects, the neural network uses a bi-directional LSTM to model the structure of a chain-like mass-spring system. While LSTMs are usually used to propagate information in time, here we use it to propagate information along the rope's mass-spring chain in both directions. Recurrently applying the same LSTM cell to each node in the rope ensures that the same physical law is applied, whether the node is closer to the endpoint or in the middle of the rope. Details of the network are described in Appendix (III). We also experimented with the recently proposed graph network \cite{InteractionNet} but found it less effective in propagating along a long chain of nodes.
When generating training data, physical parameters used in the simulation are identified automatically using CEM on a small set of real data. Simulation sequences with random actions are generated and the model is trained on one-step prediction.

\subsection{Rope manipulation with MPC}
\label{sec:mpc}
We use model predictive control to plan for a sequence of actions that takes the rope from the current configuration to the goal configuration. Both are estimated from input images. We formulate actions as first selecting a grasping point on the rope, and then selecting a 2D planar vector to move the gripper and the rope being grasped. This is different from the action space used in previous works \cite{DVF_journal, ZSVI}, where a grasping point is selected in image space, and a large portion of the action space will not make contact with the rope. Note that if our estimated rope configuration deviates from the real rope significantly, the robot may still fail to grasp the real rope. However, such cases rarely happen in our experiments, since minor errors from the perception network can be corrected in the refinement process with our proposed loss. 

\comment{
\begin{algorithm}[!t]
\caption{Algorithm for planning actions.}
\label{alg:mppi}
\small
\begin{algorithmic}
\STATE $I_{\text{goal}} \gets \text{goal image}$
\STATE $ \text{state } s_{\text{goal}} \gets \text{perception and refine}(I_{\text{goal}})$

\FOR {$t = 1,\dots,T$}
    \STATE $I \gets \text{observed image}$
    \STATE $\text{state } s \gets \text{perception and refine}(I)$

    \FOR {\text{each candidate node on rope}}
        \STATE $\text{Generate candidate action sequences } u_k$
        \STATE $\text{Rollout with dynamics model }  s_k=\text{dynamics}(s,u_k)$
        \STATE $\text{Evaluate loss } l_k=\|s_k-s_{\text{goal}}\|$
        \STATE $\text{Aggregate candidate sequences:}$
        \STATE $~~~~~~ w_k=\exp(-\alpha l_k)$
        \STATE $~~~~~~ u = \sum_k(w_k u_k) / \sum_k w_k$
        \STATE $\text{Rollout with dynamics model } s_{\text{node}}=\text{dynamics}(s,u)$
        \STATE $\text{Evaluate loss } l_{\text{node}}=\|s_{\text{node}}-s_{\text{goal}}\|$
    \ENDFOR
    \STATE $\text{Find node with minimal } l_{\text{node}}$
    \STATE $\text{Execute first few actions from corresponding } u_{\text{node}}$
\ENDFOR
\end{algorithmic}
\end{algorithm}
}

We adapt a sampling-based approach, MPPI \cite{MPPI}, for planning actions to manipulate the rope. 
We perform a nested optimization to obtain the grasping points as well as movement trajectories. In the inner loop, we sample $n$ movement trajectories of a given grasping point on the rope. These trajectories are rolled-out with our dynamics model over a time horizon $T$. The cost of each rope state along the trajectory is its distance to the goal state. 
The optimal trajectory per grasping points is computed as the cost-weighted average of the sampled trajectories, derived in \cite{MPPI}. For the outer loop, we sample grasping points on the rope and run the inner optimization loop for each in parallel. The grasping point with lowest cost of its optimal trajectory is selected.

Because the explicit rope states are available, defining an informative loss as well as sampling promising action candidates for MPPI is straight forward, compared to methods that operate in image space \cite{DVF_journal}. See Appendix (IV) for more details.

%% file: result.tex
We evaluate each of the components described in the above section, and demonstrate that both our perception and dynamics model can be trained more effectively compared to baseline models that do not incorporate any prior structure. Our proposed self-supervising learning objective is able to transfer the perception model from simple rendered images to real images with unseen occlusions. In addition, the components work with each other to achieve efficient manipulation of ropes to match visually specified goals, both in simulation and on real robots.

\subsection{Perception networks comparison}

\begin{table}
\vspace*{0.08in}
\renewcommand{\arraystretch}{1.3}
\caption{Estimation accuracy of perception networks. We report root mean square of the Euclidean distance (in meters) between estimated and ground truth point positions on the rope.}
\label{tab:label1}
\centering
\footnotesize
    \begin{tabular}{c|c|c}
    \hline
         &  Train & Test \\ \hline
    Baseline: Direct Estimate & 0.0104 & 0.0354 \\ \hline
    Ours: Coarse-to-Fine & 0.0177 & \textbf{0.0231} \\ \hline
    \end{tabular}
\end{table}

We evaluate estimation accuracy and generalization ability for two CNNs. The baseline model directly outputs 65 point coordinates from the last fully connected layer. We compare this to our proposed network that uses STNs for a coarse-to-fine estimation. Both models are trained on 10000 rendered images of b-spline curves. We report the training and evaluation loss for each network in Table~\ref{tab:label1}. Although the training loss for our network is larger than that of the baseline, our network achieves 30\% less error on a held out test set, demonstrating better generalization due to our coarse-to-fine formulation. 

Because the state space of a rope is very high dimensional, densely sampling in this space is difficult and would lead to a data set whose size is exponential in the number of rope points. Therefore, generalization as provided by the hierarchical STNs is very important for our problem.

\subsection{Self-supervised finetuning}

\label{subsec:evalimageloss}
\begin{figure}
    \centering
    \vspace*{0.08in}
    \includegraphics[width=0.35\textwidth]{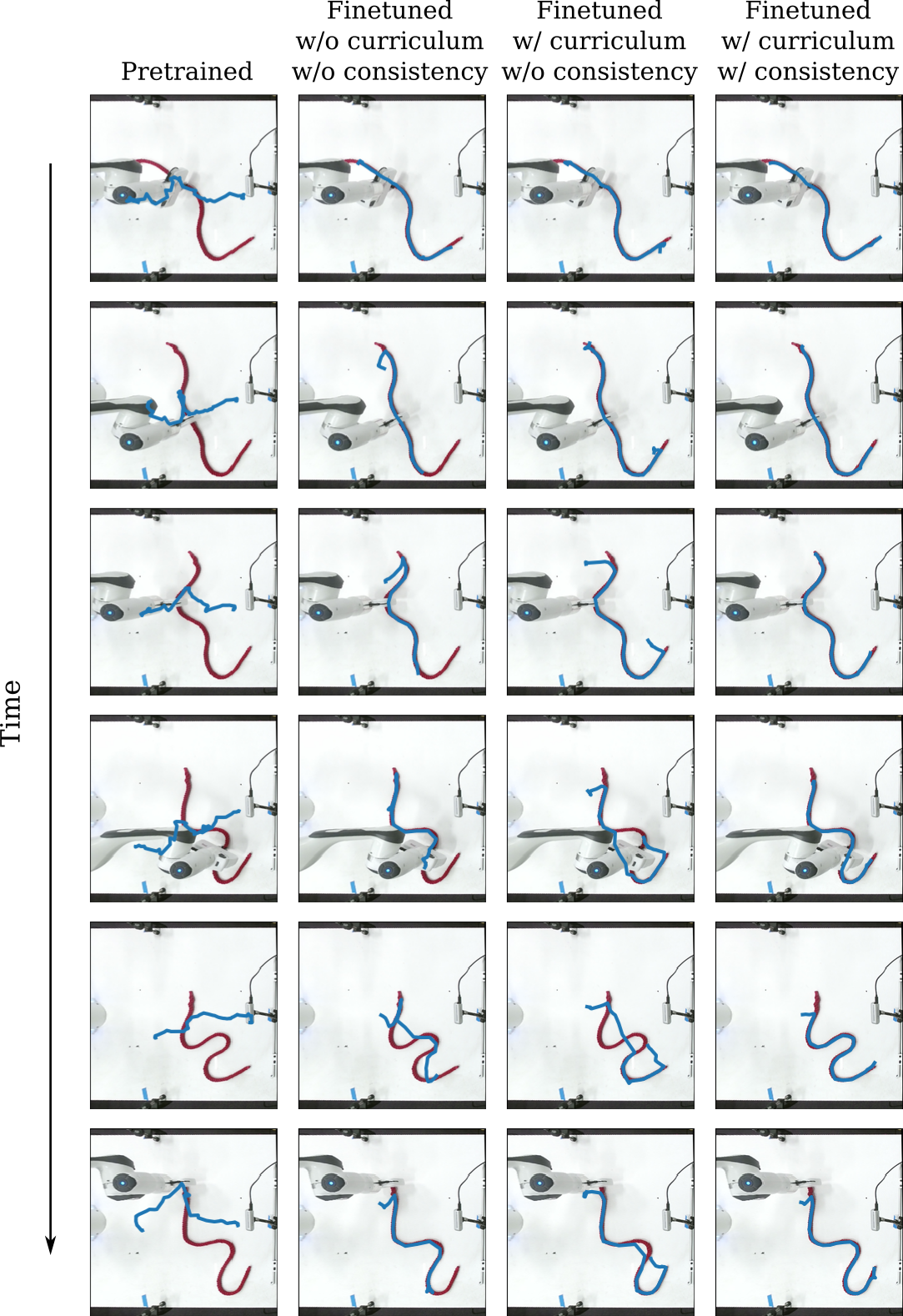}
    \caption{Visualizations of finetuning the perception network weights with the proposed objective. First column: input images overlaid with state estimation before finetuning (in blue). Second column: model finetuned with the proposed objective, without curriculum learning or temporal consistency. Third column: model finetuned with curriculum learning but no temporal consistency. Fourth column: model finetuned with both curriculum and temporal consistency. The selected samples are from a training sequence.}
    \label{fig:finetuning_image_loss}
\end{figure}

Since the robot arm is not modeled in the renderer, a network only trained on rendered images never sees the robot arm or the resulting occlusion of the rope, and thus it does not generalize well to real images (see Fig.~\ref{fig:finetuning_image_loss}). We use our proposed learning objective to finetune the parameters of our perception network on $5122$ real images, without requiring annotations. We visualize the result after finetuning in Fig.~\ref{fig:finetuning_image_loss}. Ablation studies confirm that both automatic curriculum learning and temporal consistency brings significant improvements. The improvement is not sensitive to the selection of loss threshold, shown in Appendix (VI) B.

\subsection{Learning dynamics models}
\label{subsec:eval_prediction}

We evaluate the long-horizon prediction accuracy of the learned dynamics model on both simulated and real data. We use two distance metrics for rope states: the average and the maximum deviation. Given a pair of rope states we first compute the Euclidean distance $d_i$ for each pair of corresponding points, $i=1,\dots,65$. The average and maximum deviation are defined as $\text{mean}(d_i)$ and $\max(d_i)$. These metrics will be used for all the following experiments.

\begin{figure}
\centering
    \includegraphics[width=0.245\textwidth]{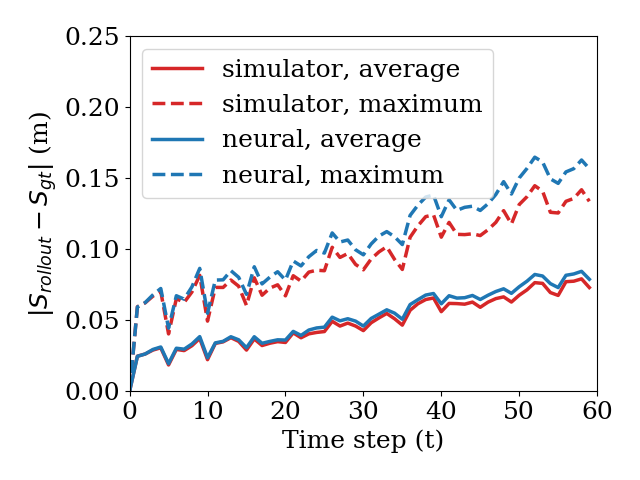}
    \hspace{-5pt}
    \includegraphics[width=0.22\textwidth]{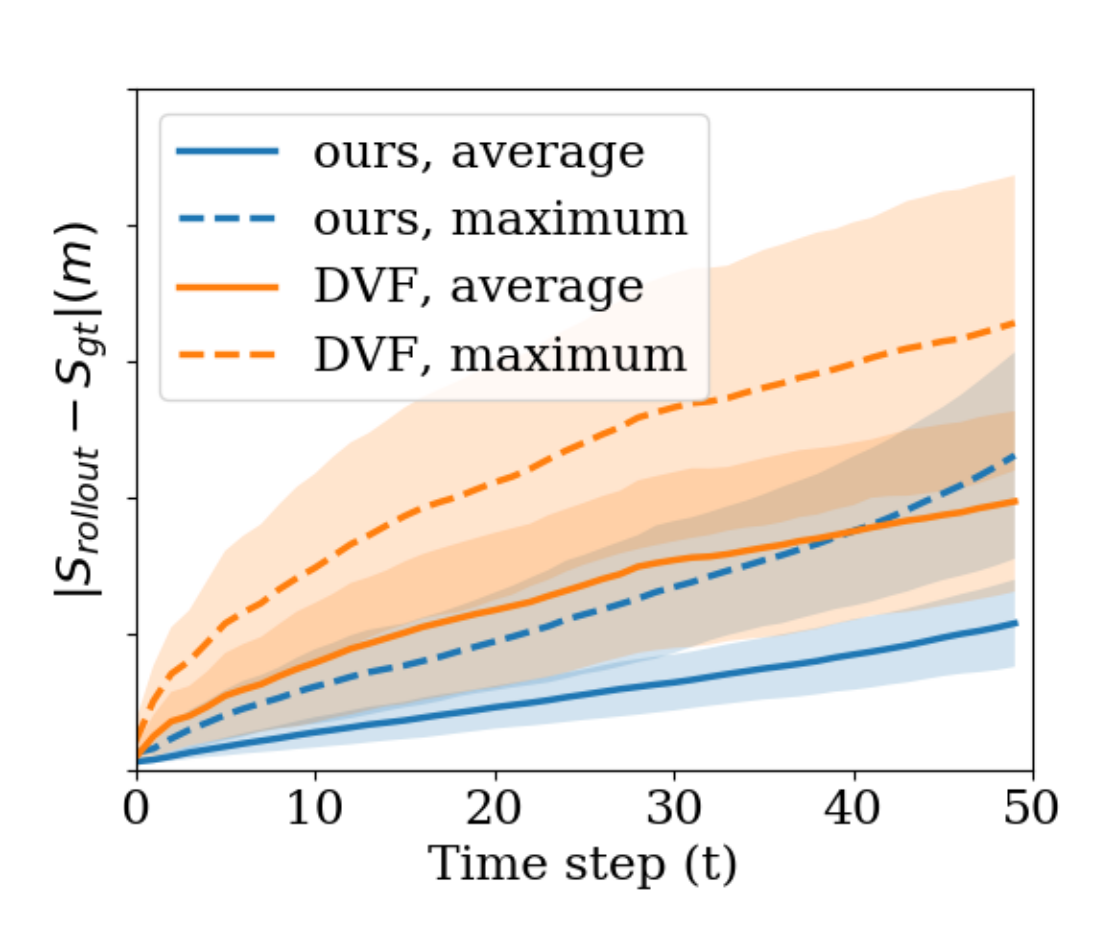}
    \caption{
    Left: average (solid lines) and maximum (dashed lines) deviations from dynamic model predictions to estimated states from real observations, averaged over $27$ sequences. Right (sharing y axis): average (solid lines) and maximum (dashed lines) deviations from dynamic model predictions to ground truth states in simulation. Lines represent the average from $200$ sequences and shaded regions represent the standard deviation.}
    \label{fig:compare_sim}
\end{figure}

\paragraph{Fast and accurate network} We show the prediction accuracy of the neural network dynamics model on real data, and compare to the simulator it is trained from. As shown in Fig.~\ref{fig:compare_sim} (left), the prediction accuracy of our neural network model is comparable to the simulation engine, with the initial state as well as ground truth states estimated from images. Noise in the lines are partially due to the noise of state estimation since the ropes are partially occluded. We expect the prediction accuracy to be further improved if also trained with multi-step prediction and finetuned on real data. The main advantage of the neural dynamics model is that it is significantly faster to predict, taking $0.03$ second per action on average, compared to $1.15$ second per action for the simulator. The neural network model is also readily parallelized on GPU with batch size up to $32000$. Both aspects are beneficial to MPC, since a lot of mental rollouts are required in parallel. 

\paragraph{Benefit of incorporating a physics prior}
\label{experiment:prediction}
We further compare the long-horizon prediction accuracy of our neural network dynamics model with the visual dynamics model (DVF) from \cite{DVF_journal}, on a batch of $200$ sequences from the simulated dataset. Both models are trained with the same dataset, except that DVF takes images, whereas our model takes explicit rope states. 
For each sequence, both models receive the same starting image and a sequence of $50$ actions. The input state for our model is estimated from the image.
For DVF, the model tracks $65$ points on the rope by predicting heat maps in image space, and point positions are the expectations from predicted point distributions.
As shown in Fig.~\ref{fig:compare_sim} (right), our neural dynamics model is significantly more accurate than DVF. Note that both models are trained on a large dataset of 0.5M simulated actions, equivalent to at least 600 robot hours. Further experiments on 1-step prediction show that our proposed model can achieve 48\% error reduction measured by maximum deviation, and 68\% error reduction measured by average deviation, when only using 3\% of the training data, compared to training the baseline model using the entire set (Appendix (VI)C). By trading off some level of generality of the method, we are able to incorporate the physics prior for deformable linear objects and achieve significant gain in data efficiency and generalization, manifested in the prediction accuracy and subsequent manipulation performance (Sec \ref{subsec:result_manip}).

\subsection{Manipulation results}

\label{subsec:result_manip}
We evaluate the performance of our system on the task of manipulating a rope on the table to match desired goal states.
We compare our method with the baseline method \cite{DVF_journal}, which uses MPC with the pixel distance cost. For the baseline, a task is specified by the start and goal positions of $11$ equidistant points on the rope. To select promising grasping points from images, we compute the pixel-wise difference between the current observed image and the goal image, and only sample grasping positions where the two images are significantly different.

\begin{figure}
    \centering
    \includegraphics[width=0.25\textwidth]{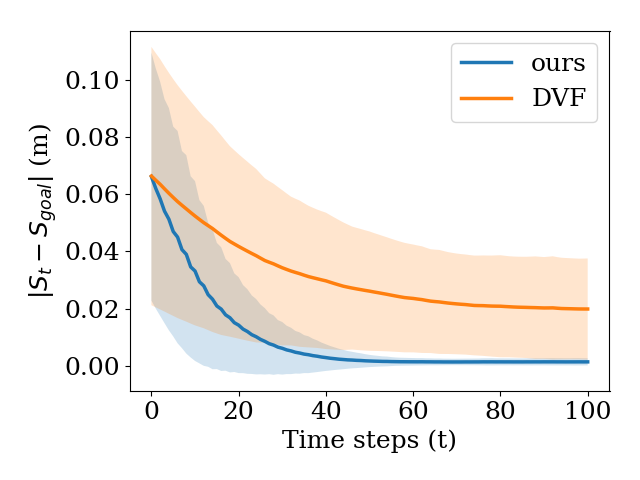}
    \caption{
    Residual average deviation between the rope state at each time $t$ and the goal state, for each manipulation method. Mean and standard deviation are obtained from $100$ experiments of random start state and goal state pairs.}
    \label{fig:manipulation_compare}
\end{figure}

For quantitative evaluation, we run manipulation tasks in simulation, and select start states and goal states from randomly generated b-spline curves. Both our method and the baseline only see the rendered images.
We report the distance to goal $L(t)$ as a function of time $t$, where $L(t)$ is the average deviation from the current rope state to the specified goal state. We show the mean and standard deviation over $100$ independent experiments in Fig.~\ref{fig:manipulation_compare}. Our method achieves the goal state within $60$ steps in most cases, and the remaining distance is very small, while the baseline method that operates in image space often cannot achieve the goal state within $100$ steps, showing large residual distances at $t=100$. For more visualizations refer to Appendix (V).


\begin{figure}
    \centering
    \includegraphics[width=0.23\textwidth]{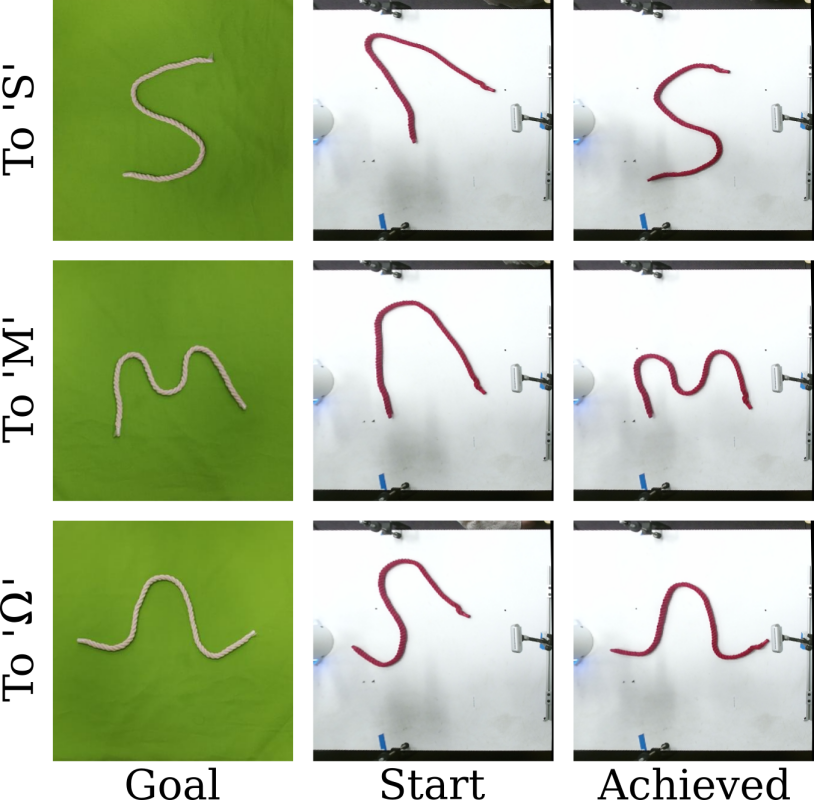}
    \caption{From left to right: the start state, goal state, and achieved state after $40$ steps. Goal images are rescaled for best comparison. Other images are taken by Kinect and projected to top-down view.}
    \label{fig:real_robot_manip}
\end{figure}

We also demonstrate rope manipulation on the real robot. We arrange the goal state of the rope to be an ``S" shape, a ``W" shape, or an ``$\Omega$" shape. Due to the very large state space of ropes, the network does not always generalize to the rope states during the manipulation tasks, which may be outside of the training distribution. Most of the overfitting is from the coarsest prediction layer, and the refinement layers generalize well if given a reasonable coarse prediction. This motivates us to further exploit temporal information during manipulation. Based on the latest estimated state $s_t$, the MPC plans the optimal action $a_t$. The learned dynamics model predicts the next state $\hat{s}_{t+1}$. $\hat{s}_{t+1}$ is subsampled into $8$ segments to feed into the CNN's first STN, together with the next image $I_{t+1}$, and the CNN refines $\hat{s}_{t+1}$ into the next estimated state $s_{t+1}$. Using this temporal information greatly improved generalization of our perception method. Appendix (VI)A shows the tracking results where an intersection is made with the rope, which is never seen during training.

To highlight one benefit of using explicit state representations, we use a different rope on a different background to demonstrate goals. The two ropes have different appearance as well as different lengths and thicknesses. The L2 loss between goal and observed images would not be informative for video prediction methods, and it would be hard to embed the goal image into the latent space of the manipulated rope, if using \cite{Abbeel_RSS19}. Our method successfully finished all 3 tasks,  visualized in Fig.~\ref{fig:real_robot_manip}. Also see supplementary material for robot manipulation videos.

%% file: conclusion.tex
We demonstrated model-based, visual robot manipulation of deformable linear objects. Our forward model makes explicit estimation of rope states from images, and learns a dynamic model in state space. We proposed a self-supervising learning objective to enable self-supervised continuous training of rope state estimation on real data, without requiring expensive annotations. Our objective is generalizable across a wide range of visual appearances. We also proposed coarse-to-fine state estimation using hierarchical STNs that greatly improves generalization to unseen rope shapes. With access to the rope's explicit state, we are able to incorporate physics priors, e.g., the structure of mass-spring systems, into the design of the network structure for dynamics models, in addition to using physical simulation for data generation. 
We demonstrated that our dynamics model using bi-directional LSTM 
has reduced prediction errors by up to 68\% while only using 3\% of total training data, compared to a baseline dynamics model in pixel space, and that our method achieves more efficient manipulation in matching visually specified goals. Although we only demonstrated manipulation on a horizontal plane, our method can be extended to 3D manipulation tasks with minor modifications. In future works, instead of assuming a known table height, we can use the estimated 2D states to extract depth values for the Kinect depth images to reproject the estimated points to 3D.

For future directions, it would be interesting to explore using our self-supervising learning objective to continuously train the perception and dynamics network while the robot is performing manipulation tasks, similar to DAGGER \cite{DAGGER}. Our neural network dynamics model can be extended to have object's physical properties as latent variables, such that the dynamics model can adapt quickly to ropes/wires with different physical properties, by updating the latent variables instead of network weights. Finally, we would also like to extend this method to deformable objects with even higher dimensional state spaces, such as clothing. Possible directions include adapting the proposed perception method to estimate the contour of clothes, and extending the proposed bi-directional LSTM to a 2-dimensional bi-directional LSTM to model the dynamics of clothes.